\documentclass[10pt,twocolumn,letterpaper]{article}

\usepackage[pagenumbers]{cvpr} 

\usepackage[dvipsnames]{xcolor}

\usepackage{graphicx}
\usepackage{amsmath}
\usepackage{amssymb}
\usepackage{booktabs}
\usepackage{makecell}
\usepackage{multirow}
\usepackage{pifont}

\definecolor{cvprblue}{rgb}{0.21,0.49,0.74}
\usepackage[pagebackref,breaklinks,colorlinks,citecolor=cvprblue]{hyperref}

\title{SkySense: A Multi-Modal Remote Sensing Foundation Model Towards Universal Interpretation for Earth Observation Imagery}

\author{
Xin Guo\textsuperscript{1}\thanks{Equally contributing first authors.\ \ $^{\dagger}$Corresponding authors.\ \ $^{\S}$Work done during the internship of the author at Ant Group.}~, Jiangwei Lao\textsuperscript{1}\footnotemark[1]~, Bo Dang$^{\S}$\textsuperscript{2}\footnotemark[1]~, \\ Yingying Zhang\textsuperscript{1}, Lei Yu\textsuperscript{1}, Lixiang Ru\textsuperscript{1}, Liheng Zhong\textsuperscript{1}, Ziyuan Huang\textsuperscript{1}, Kang Wu$^{\S}$\textsuperscript{2}, Dingxiang Hu\textsuperscript{3,1}, \\ Huimei He\textsuperscript{3,1}, Jian Wang\textsuperscript{1}, Jingdong Chen\textsuperscript{1}, Ming Yang\textsuperscript{1}$^{\dagger}$, Yongjun Zhang\textsuperscript{2}, Yansheng Li\textsuperscript{2}$^{\dagger}$\\
\textsuperscript{1}Ant Group \quad 
\textsuperscript{2}Wuhan University \quad
\textsuperscript{3}MYBank \\
{\tt\small \{bangzhu.gx, wenshuo.ljw\}@antgroup.com}, {\tt\small bodang@whu.edu.cn}}

\begin{document}
\maketitle
\begin{abstract}
Prior studies on Remote Sensing Foundation Model (RSFM) reveal immense potential towards a generic model for Earth Observation. Nevertheless, these works primarily focus on a single modality without temporal and geo-context modeling, hampering their capabilities for diverse tasks. In this study, we present SkySense, a generic billion-scale model, pre-trained on a curated multi-modal Remote Sensing Imagery (RSI) dataset with 21.5 million temporal sequences. SkySense incorporates a factorized multi-modal spatiotemporal encoder taking temporal sequences of optical and Synthetic Aperture Radar (SAR) data as input. This encoder is pre-trained by our proposed Multi-Granularity Contrastive Learning to learn representations across different modal and spatial granularities. To further enhance the RSI representations by the geo-context clue, we introduce Geo-Context Prototype Learning to learn region-aware prototypes upon RSI's multi-modal spatiotemporal features. To our best knowledge, SkySense is the largest Multi-Modal RSFM to date, whose modules can be flexibly combined or used individually to accommodate various tasks. It demonstrates remarkable generalization capabilities on a thorough evaluation encompassing 16 datasets over 7 tasks, from single- to multi-modal, static to temporal, and classification to localization. SkySense surpasses 18 recent RSFMs in all test scenarios. Specifically, it outperforms the latest models such as GFM, SatLas and Scale-MAE by a large margin, i.e., 2.76\%, 3.67\% and 3.61\% on average respectively. We will release the pre-trained weights to facilitate future research and Earth Observation applications.
\end{abstract}
\vspace{-1.2em}
\section{Introduction}
\label{sec:intro}

\begin{figure}
	\centering
		\includegraphics[scale=.35]{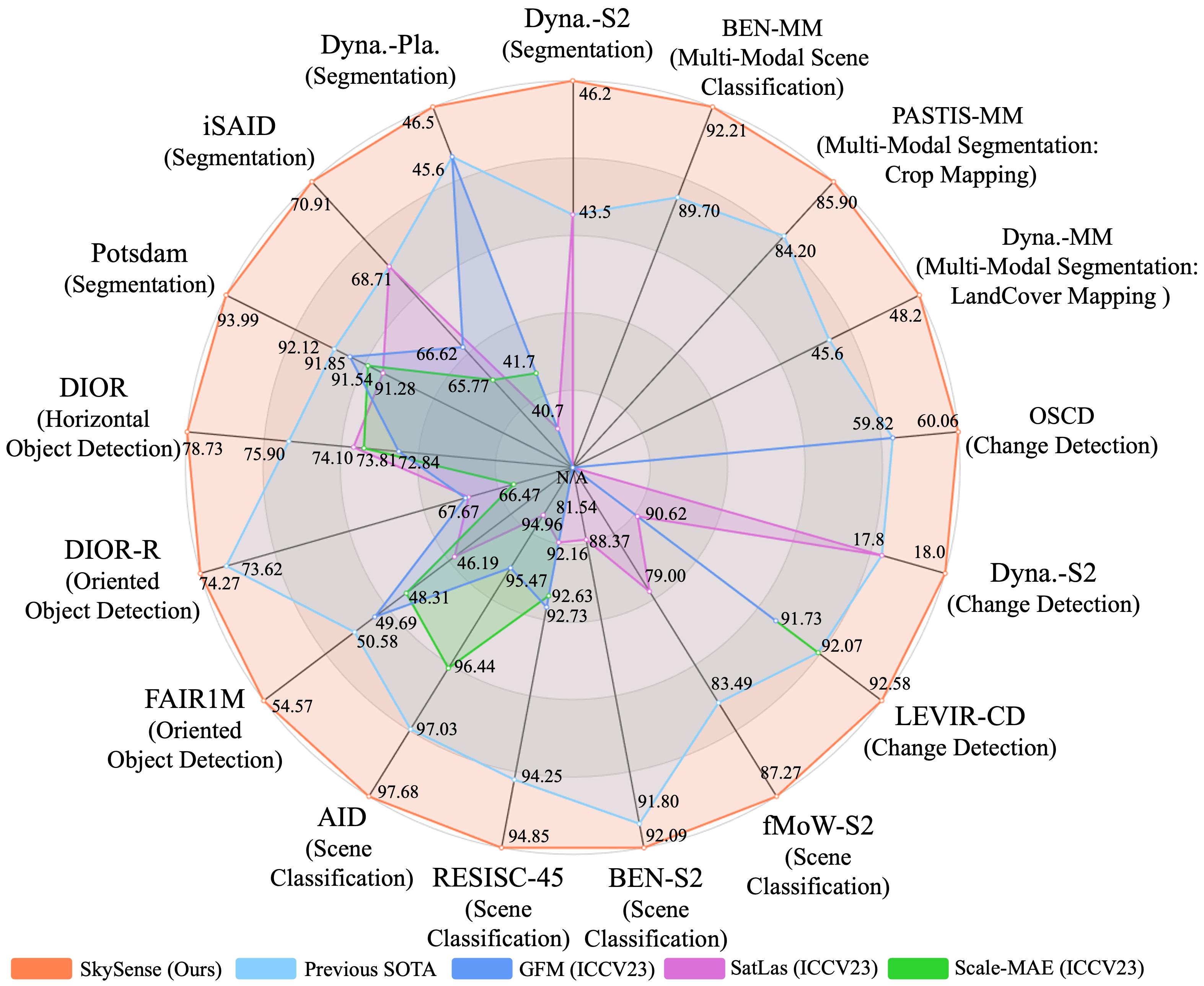}
	    \caption{SkySense has achieved superior performance on 16 datasets over 7 distinct tasks compared with 18 state-of-the-art RSFMs and supports a board range of EO imagery interpretations.}
    \vspace{-1.2em}
	\label{fig:fig1}
\end{figure}

Remote Sensing Imagery (RSI) interpretation is crucial in understanding our common home, the Earth~\cite{chi2016big,tuia2023artificial}, via quite diverse tasks~\cite{yuan2021review,cheng2016survey,lv2022land,cao2023multi}, \eg crop monitoring, natural disaster management, \etc Every task may require significant dedicated efforts and resources to build a task-specific model. Recently, Foundation Model emerges as a pre-trained generic model that excels in a wide range of downstream tasks~\cite{zhou2023comprehensive,yang2022continual}. Hence, there is a soaring interest in exploring a comprehensive Remote Sensing Foundation Model (RSFM) for many Earth Observation (EO) tasks.

The key question naturally arises: \emph{What is essential for a RSFM?} First of all, an ideal RSFM should possess the ability to perceive multi-modal temporal RSI. EO heavily relies on multi-modal time series of remote sensing data, including temporal optical and Synthetic Aperture Radar (SAR) data. Individual modality offers unique advantages and complements to each other. For example, optical images provide rich spectral bands and texture details but are susceptible to weather~\cite{zorzi2022polyworld}. In contrast, SAR sensors capture clear imagery in all weather conditions~\cite{huang2017opensarship,li2022mcanet}. Moreover, the time series of such data provide the crucial temporal clue to various tasks~\cite{garnot2022multi,cao2023multi,zhang2023demonstration} like change prediction. Second, a RSFM should be easy to tailor when being deployed for EO tasks using different modalities (\ie, single- and multi-modal) at different spatial (\ie, pixel-, object-, and image-level) granularities. Last but not the least, remote sensing data is inherently contingent on their space-time coordinates, which provide rich regional and seasonal geo-context that benefits RSI interpretation a lot, as indicated in~\cite{liu2023seeing,chen2019collaborative,guo2022isdnet,huang2022toward,mfvnet}. Therefore, a RSFM shall bear the vital capability of effective geo-context learning and utilization.

\begin{table}
\centering
\setlength\tabcolsep{1.5pt}%
\scriptsize
\begin{tabular}{l|cccc}
\toprule
\multirow{2}{*}{Model} & \multicolumn{4}{c}{Different EO Interpretation Input Types}\\ \cmidrule(lr){2-5}
                       & \begin{tabular}[c]{@{}c@{}}Single-Modal\\O(RGB)\end{tabular} & \begin{tabular}[c]{@{}c@{}}Single-Modal\\O(Ms)\end{tabular} & \begin{tabular}[c]{@{}c@{}}Multi-Modal\\Static O \& SAR\end{tabular} & \begin{tabular}[c]{@{}c@{}}Multi-Modal\\Temporal O \& SAR\end{tabular} \\ \midrule
SkySense               & \ding{52}                                                 & \ding{52}                                                  & \ding{52}                                                     & \ding{52}                                                        \\ \midrule
SatLas\cite{bastani2022satlas}                   & \ding{52}                                                   & \ding{52}                                                  &                                                      &                                                        \\
GFM\cite{mendieta2023gfm}                   & \ding{52}                                                   &                                                  &                                                      &                                                        \\
Scale-MAE\cite{reed2022scale}                 & \ding{52}                                                   &                                                 &                                                      &                                                        \\ \bottomrule
\end{tabular}
\caption{SkySense supports various input types. O(RGB): Optical RGB images; O(Ms): Optical multispectral images.}
\vspace{-1.2em}
\label{tab1}
\end{table}

Previous works on RSFM~\cite{ayush2021geography, manas2021seasonal, akiva2022self, wanyan2023dino, cong2022satmae, sun2022ringmo, wang2022advancing, cha2023billion, tao2023tov, wang2022ssl4eo, muhtar2023cmid, mall2023change, bastani2022satlas, reed2022scale, mendieta2023gfm, wang2023scaling, jakubik2023foundation} have demonstrated their preliminary success on several specific datasets. However, these RSFMs, while proficient in certain areas, are limited in their applications to EO tasks, due to factors such as single-modal pre-training and the neglect of geo-context.

In this paper, we propose SkySense, a billion-scale Multi-Modal Remote Sensing Foundation Model (MM-RSFM). SkySense incorporates 2.06 billion parameters and is pre-trained on a large-scale multi-modal dataset which comprises 21.5 million RSI temporal sequences extracted from high-spatial-resolution optical images (HSROIs), medium-resolution temporal multispectral imagery (TMsI) and temporal SAR imagery (TSARI). To handle the multi-modal temporal RSI sequences, SkySense employs a factorized multi-modal spatiotemporal encoder to perform spatial feature extraction and multi-modal temporal fusion independently, since RSI sequence are spatially-aligned in nature. It leads to a modular design allowing flexible use of its modules, \ie, the spatial encoder can be either used alone or in combination of the fusion module to support tasks from static single-modal to temporal multi-modal. This design delivers strong modeling of RSI sequences while using substantially less parameters compared to common 3D structures~\cite{zheng20213d,m2019semantic}. The factorized encoder is pre-trained by Multi-Granularity Contrastive Learning to construct features from different modal and spatial granularities. Furthermore, we propose Geo-Context Prototype Learning to generate regional prototypes from RSI features given geo-locations. This approach enhances multi-modal spatiotemporal representation learning by leveraging the regional context clue hidden in numerous unlabeled RSI.

SkySense has achieved the state-of-the-art (SOTA) performance across a variety of modalities and EO tasks, as shown in \cref{fig:fig1}. We evaluate SkySense on a diverse set of 16 datasets \cite{waqas2019isaid,toker2022dynamicearthnet,li2020object,cheng2022anchor, sun2022fair1m,xia2017aid,cheng2017remote,sumbul2020bigearthnet,cong2022satmae,chen2020spatial,daudt2018urban,garnot2022multi}, where the selection covers different task types, modalities and spatial scales. The results demonstrate that SkySense outperforms 18 advanced RSFMs \cite{ayush2021geography, manas2021seasonal, akiva2022self, wanyan2023dino, cong2022satmae, sun2022ringmo, wang2022advancing, cha2023billion, tao2023tov, wang2022ssl4eo, muhtar2023cmid, mall2023change, bastani2022satlas, reed2022scale, mendieta2023gfm, wang2023scaling} in all test scenarios, validating its competitive edge for a broad range of EO interpretation tasks. \cref{tab1} compares our work with latest representative studies w.r.t. various input types of EO interpretation.

In summary, our technical contributions are:
\begin{itemize}
\item We propose SkySense, the largest MM-RSFM to date with a modular design, which is capable of handling diverse tasks, from single- to multi-modal, static to temporal, and classification to localization.
\item The design of SkySense involves three novel technical components: a) A factorized multi-modal spatiotemporal encoder to effectively process multi-modal temporal RSI; b) Multi-Granularity Contrastive Learning that learns features at various levels of granularities to facilitate different tasks; c) Geo-Context Prototype Learning to extract region-aware geo-context clue to enable implicit geo-knowledge integration.
\item We extensively compare SkySense with 18 recently published RSFMs. Our model has achieved the SOTA performance, supparssing the latest models like GFM, SatLas and Scale-MAE by over $2.5\%$ on average. We hope the release of pre-trained weights will contribute to the Remote Sensing community and facilitate future research.
\end{itemize}



\section{Related Work}


\subsection{Remote Sensing Foundation Model}
Recent Remote Sensing Foundation Models draw their primary inspiration from the research on Vision Foundation Model~\cite{he2016deep, vaswani2017attention,liu2021swin,deng2009imagenet, he2020momentum,chen2020simple,grill2020bootstrap,radford2021learning,caron2021emerging,he2022masked,bao2022beit}. Remote sensing data inherently integrates space-time coordinates and has diverse spatial scales. The maintstream RSFMs extend the foundation model techniques to space-time RS data, such as Contrastive Learning. For instance, GASSL \cite{ayush2021geography} utilized geo-location prediction as an additional pre-text task in the MoCo-v2 framework \cite{chen2020mocov2}. Multiple views with different sizes were utilized by DINO-MC \cite{wanyan2023dino} for self-supervised learning within the DINO framework \cite{caron2021emerging}. SeCo \cite{manas2021seasonal} and CACo \cite{mall2023change} both proposed Contrastive Learning to perceive short-term and long-term changes by using the spatiotemporal structure of temporal RSI sequences. Besides, there are works either improving the MIM-based framework \cite{sun2022ringmo, wang2022advancing, reed2022scale} or exploring the model scale-up \cite{cha2023billion}. For example, RingMo \cite{sun2022ringmo} modified MAE to adapt to the dense objects in RSI. SatMAE \cite{cong2022satmae} employed TMsI to enhance the performance on temporal sequences. Scale-MAE \cite{reed2022scale} built a framework with scale-aware encoder. Recent efforts such as CMID \cite{muhtar2023cmid} and GFM \cite{mendieta2023gfm} have commenced to explore amalgamation of CL and MIM strategies. Concurrently, CROMA \cite{fuller2023croma} and DeCUR \cite{wang2023decur} investigated multi-modal pre-training for single- and multi-modal tasks using static imagery. In this study, we propose a comprehensive MM-RSFM, SkySense, to fill the gap in existing RSFMs, \ie, single modality of RingMo, CACo, \etc, static input of Scale-MAE, CROMA, \etc, and the neglect of geo-context of SatLas, RVSA, \etc 

\begin{figure}
	\centering
		\includegraphics[scale=.47]{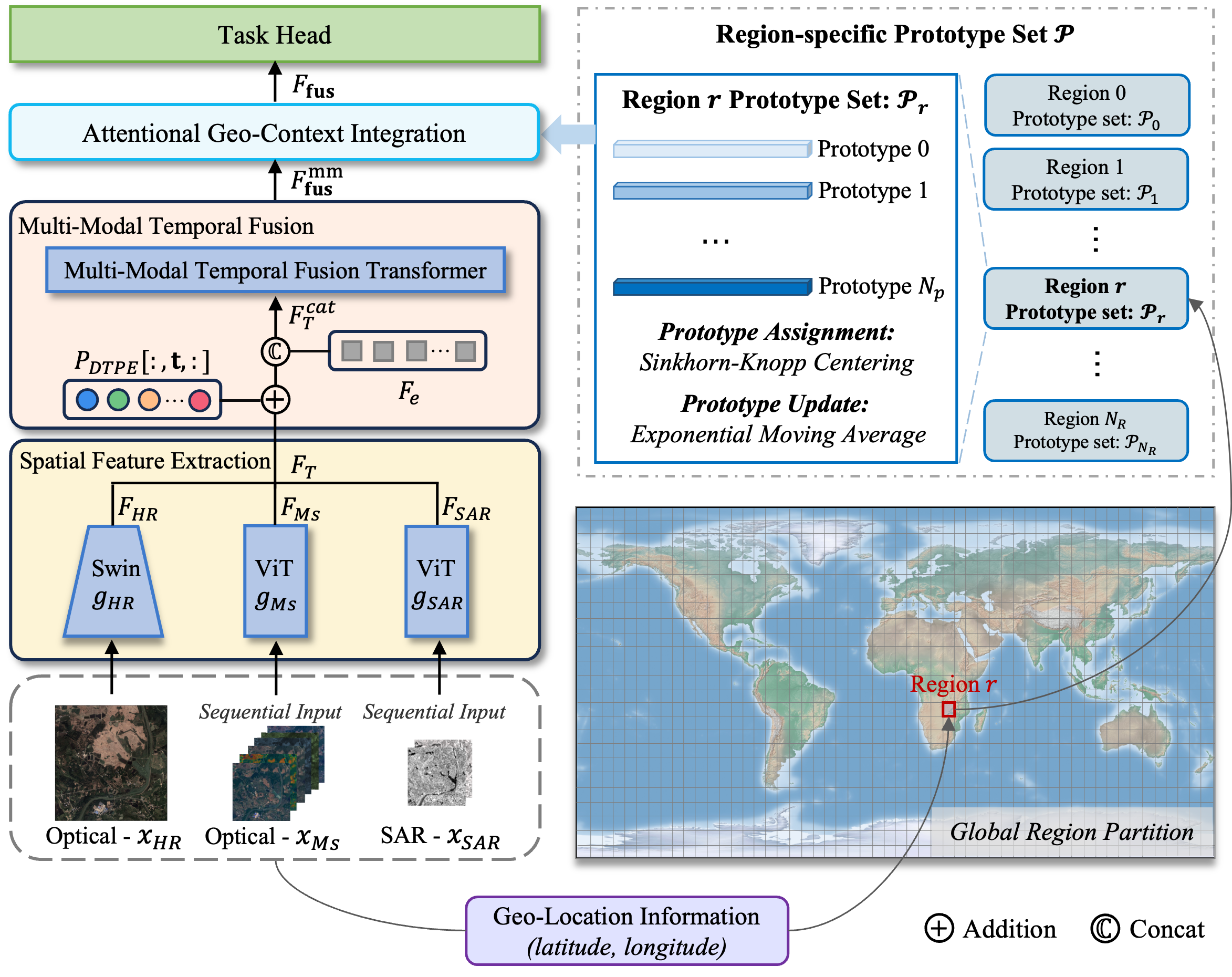}
	    \caption{The overview of our SkySense model architecture.}
    \vspace{-1.2em}
	\label{fig:fig2}
\end{figure}
\section{SkySense}
In this section, we introduce the pre-training dataset and the design choices for individual module respectively.

\subsection{Pre-training Dataset}
We curate an extensive multi-modal remote sensing dataset with temporal sequences, containing RSI from various sources: HSROIs from WorldView-3, 4, \etc (RGB band), TMsI from Sentinel-2 (B2-8, B8A, B11-12 band) and TSARI from Sentinel-1 (VV, VH Polarization). All data is geo-spatially aligned. Strictly speaking, HSROIs and TMsI shall be categorized to the optical modality, while TSARI falls to the SAR modality. However, due to HSROIs and TMsI's significant difference in spectral band and ground sample distance, we regard HSROIs and TMsI as two distinct modalities for simplicity in this paper. The dataset comprises 21.5 million training samples, each consisting of a static HSROI with rich texture details, a TMsI containing temporal and multispectral data, a TSARI providing backscatter polarization under cloud coverage, and the metadata like geo-location and acquisition date for geo-context modeling. This dataset covers a great variety of scenarios across resolution, spectrum, and imaging mechanism. More details of the data are included in the supplementary materials. We construct the input for SkySense as $\left\{x_{HR}, x_{Ms}, x_{SAR}\right\}$, where $x_{HR}$ represents a static HSROI; $x_{Ms}$ is a Sentinel-2 TMsI after filtering cloudy images, where we randomly select 20 images to form the sequence; and $x_{SAR}$ stands for a standard-calibrated TSARI, from which we randomly select 10 images for training.

\subsection{Model Architecture}\label{modelarch}
\noindent{\textbf{Factorized Multi-Modal Spatiotemporal Encoder.}}
The overall architecture of our method is illustrated in \cref{fig:fig2}. In a multi-modal input $\left\{x_{HR}, x_{Ms}, x_{SAR}\right\}$, the pixels within each RSI naturally align with the others given the same geo-location. Upon this, we propose a factorized encoder that initially extracts spatial features from each RSI independently and then fuses them to capture a multi-modal spatiotemporal representation. The design separates spatial feature extraction from the feature fusion, enabling the integration of the clues from modality, time and geo-context.

\textit{Spatial Feature Extraction.} To handle the spatially aligned sequence input $\left\{x_{HR}, x_{Ms}, x_{SAR}\right\}$, we utilize the spatial encoder $g_{HR}$, $g_{Ms}$ and $g_{SAR}$ for each individual RSI from HSROI, TMsI and TSARI respectively. As shown in \cref{equ:h_i}, the obtained feature $F_{i} \in \mathbb{R}^{h \times w \times T_{i} \times d}, i \in \left\{HR, Ms, SAR\right\}$ are of the same size in spatial dimension, where $h$ and $w$ are the height and width of $F_{i}$, $T_{HR}$, $T_{Ms}$, $T_{SAR}$ represent the sequence lengths of HSROI, TMsI, and TSARI respectively, and $d$ is the feature dimension. The initial multi-modal temporal feature representation $F_{T} \in \mathbb{R}^{N_S \times N_T \times d}$ is generated by concatenating all $F_{i}$ along the time dimension, where $N_{S}= h \times w$ represents the feature size in the spatial dimension, and $N_{T} = \sum_{i \in \left\{HR, Ms, SAR\right\}}T_{i}$ represents the total sequence length across all modalities,
\vspace{-0.3em}
\begin{align}
\label{equ:h_i}
\begin{gathered}
F_{i} = g_i(x_i), i \in \left\{HR, Ms, SAR\right\}, \\
F_{T} = \operatorname{Concat}\left[F_{HR}, F_{Ms}, F_{SAR}\right].
\end{gathered}
\end{align}

\textit{Multi-modal Temporal Fusion.} Next, we incorporate the date-specific temporal positional encoding $P_{DTPE}[:,\textbf{t},:] \in \mathbb{R}^{1 \times N_T \times d}$ to $F_{T}$ through broadcasting, creating $F_{T}^{date}$ for date-aware modeling. $F_{T}^{date}$ is then concatenated with an extra token $F_{\mathbf{e}} \in \mathbb{R}^{N_S \times 1 \times d}$~\cite{dosovitskiy2020image} (see \cref{deqn_ex1}),
\vspace{-0.6em}
\begin{align}
\label{deqn_ex1}
\begin{gathered}
F_{T}^{date} = F_{T}+P_{DTPE}[:,\textbf{t},:], \\
F_{T}^{cat}=\operatorname{Concat}\left[{F}_{\mathbf{e}}, F_{T}^{date}\right] \in \mathbb{R}^{N_S \times (1+N_T) \times d},
\end{gathered}
\end{align}
where $\textbf{t} \in \mathbb{R}^{N_T}$ is a vector containing the acquisition dates of all RSI in the current batch. $P_{DTPE} \in \mathbb{R}^{1 \times 365 \times d}$ is a learnable parameter representing different dates of a year, which is essential for tasks affected by seasons (\eg, crop recognition). $F_{T}^{cat}$ is then fed into the Multi-modal Temporal Fusion Transformer, composed of multiple Naive Transformer encoder layers. This module employs self-attention to integrate multi-modal temporal data, generating the multi-modal spatiotemporal feature $F_{\mathbf{fus}}^{\text{mm}} \in \mathbb{R}^{N_S \times 1 \times d}$.

\noindent{\textbf{{Attentional Geo-Context Integration.}} Each RSI's geographical location may reveal rich region-specific geo-context. It is valuable for RSI interpretation as indicated by~\cite{liu2023seeing,chen2019collaborative,guo2022isdnet,huang2022toward}. To utilize this contextual clue to enhance $F_{\mathbf{fus}}^{\text{mm}}$, we employ a region-specific prototype set $\mathcal{P} \in \mathbb{R}^{N_{R} \times N_{p} \times d}$ (shown on the right side of \cref{fig:fig2}), where $N_{R}$ is the number of regions, $N_{p}$ represents the number of prototypes for each region and $d$ denotes the feature dimension. The learning procedure of $ \mathcal{P} $ will be elaborated in \cref{pretraining}. Specifically, a regional prototype subset $\mathcal{P}_r \in \mathbb{R}^{ N_{p} \times d}$ is chosen from $\mathcal{P}$ based on the geo-location embedded with $F_{\mathbf{fus}}^{\text{mm}}$. $F_{\mathbf{fus}}^{\text{mm}}$ is then attended to the prototypes of $ \mathcal{P}_{r} $ through the attention mechanism, as shown in \cref{deqn_ex4}. The weights, computed from \( \operatorname{Softmax}\left(\frac{QK^T}{\sqrt{d}}\right) \), facilitate a soft selection of prototypes in accordance with their similarity to \( F_{\mathbf{fus}}^{\text{mm}} \). The final representation $F_{\mathbf{fus}} \in \mathbb{R}^{N_S \times 2d}$ is generated by concatenation of $F_{\mathbf{fus}}^{\text{mm}}$ and weighted sum of prototypes from $\mathcal{P}_{r}$ along the feature dimension. The prototypes represent a set of discriminative features linked to certain semantics like water body, cropland, \etc By finding the similar ones to $F_{\mathbf{fus}}^{\text{mm}}$, we provide the standard representations of certain semantics to complement $F_{\mathbf{fus}}^{\text{mm}}$,
\vspace{-0.3em}
\begin{align}
\label{deqn_ex4}
\begin{gathered}
F_{\mathbf{fus}} =\operatorname{Concat}\left[F_{\mathbf{fus}}^{\text{mm}}, \operatorname{Softmax}\left(\frac{QK^T}{\sqrt{d}}\right) V\right], \\
Q=F_{\mathbf{fus}}^{\text{mm}}, K=V=\mathcal{P}_{r}.
\end{gathered}
\end{align}

\begin{figure*}
	\centering
		\includegraphics[scale=0.053]{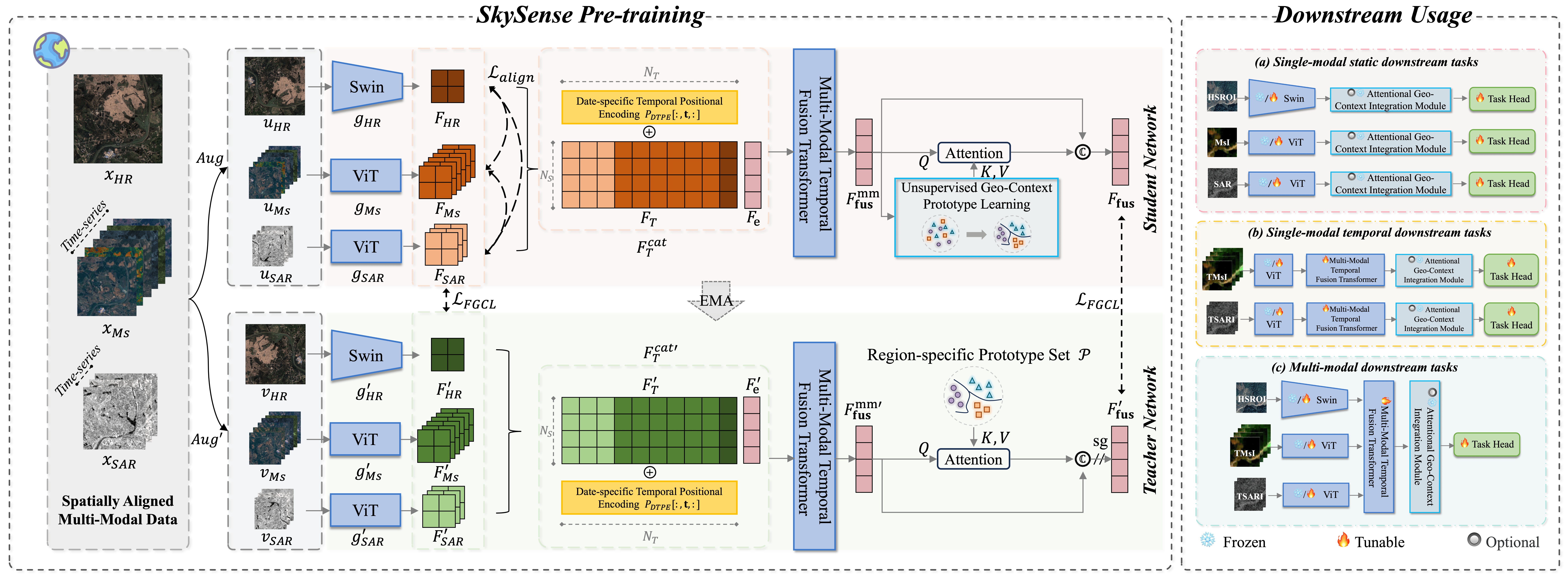}
            \vspace{-2em}
	    \caption{Overview of SkySense pre-training and downstream usage. SkySense employs data augmentations on the input and then feeds the augmented data into the student and teacher networks respectively. Multi-Granularity Contrastive Learning and Cross-Modal Alignment are proposed to pre-train the overall network. The region-specific prototype set $\mathcal{P}$ is learned on the student branch and it is frozen for downstream usage. Enhancing feature with $\mathcal{P}$ is optional. After pre-training, we adopt the parameters of the teacher branch for downstream tasks. Each pre-trained module can be used alone or combined with the others, with the chosen ones either frozen or fine-tuned.}
    \vspace{-1.2em}
	\label{fig:fig3}
\end{figure*}

\subsection{Pre-training}\label{pretraining}
An overview of our pre-training procedure is illustrated in \cref{fig:fig3}. We build the pre-training framework on a common teacher-student structure~\cite{caron2021emerging}, since it conducts self-supervised learning using only positive pairs, which is easily accessible given spatially aligned RSI, avoiding complicated design of negative pairs. Teacher's parameter set $\theta^{\prime}$ is updated through exponential moving average (EMA)~\cite{he2020momentum} from student's parameter set $\theta$. 

\noindent\textbf{Multi-Granularity Contrastive Learning.} We propose Multi-Granularity Contrastive Learning for self-supervised learning on different modal and spatial granularities for diverse tasks. Given the input $\left\{x_{HR}, x_{Ms}, x_{SAR}\right\}$, two sets of random augmentations are employed, generating two groups of views $\left\{u_{i}\right\}$ and $\left\{v_{i}\right\}$, where $i\in \{HR, Ms, SAR\}$. $u_i$ and $v_i$ are subsequently fed into the spatial encoders from the student and teacher branches respectively. $g_i$ is the student's spatial encoder and $g_i^{\prime}$ is the teacher's. The features are generated as in \cref{backbone_ema},
\vspace{-0.4em}
\begin{equation}
F_i=g_i\left(u_i\right), F_i^{\prime}=g_i^{\prime}\left(v_i\right) \quad i \in\{HR, Ms, SAR\}.
\label{backbone_ema}
\end{equation}

After applying the multi-modal temporal fusion and geo-context integration on $F_i$ and $F_i^{\prime}$, the final feature $F_{\mathbf{fus}}$ and $F_{\mathbf{fus}}^{\prime}$ are obtained. Initially, we establish pixel-, object- and image-level contrastive learning to progressively learn coarse-to-fine spatial features for various tasks.

Each temporal slice of $F_i$ can be viewed as a pixel-level feature $F_i^{\mathbf{pix}} \in \mathbb{R}^{N_S \times d}$. Pixel-level contrastive learning loss $\mathcal{L}_{\mathbf{pix}}$ is obtained by averaging all $\mathcal{L}_{CL}$ across the spatial ($s$) and temporal ($t$) dimensions, as shown in \cref{equ:pixel}. $f_i^{\mathbf{pix}} \in \mathbb{R}^{d}$ represents a feature vector from $F_i^{\mathbf{pix}}$ and $f_i^{\mathbf{pix}\prime}$ is its correspondence at the same geo-location. $\mathcal{L}_{CL}$ denotes the learning loss~\cite{caron2021emerging} between $f_i^{\mathbf{pix}}$ and $f_i^{\mathbf{pix}\prime}$, and
\vspace{-0.3em}
\begin{equation}
\begin{aligned}
\mathcal{L}_{\mathbf{pix}}(F_{i}, F_{i}^{\prime}) = \frac{1}{N_S T_i}\sum_s \sum_t \mathcal{L}_{CL}(f_i^{\mathbf{pix}}, f_i^{\mathbf{pix}\prime}).
\end{aligned}
\label{equ:pixel}
\end{equation}

$F_i^{\mathbf{obj}} \in \mathbb{R}^{N_C \times d}$ denotes object-level feature generated from unsupervised clustering on pixel-level feature vectors $f_i^{\mathbf{pix}}$ in a single RSI, where $N_C$ is the number of clusters. The clustering employs the same Sinkhorn-Knopp algorithm~\cite{caron2020unsupervised} we apply for Geo-Context Prototype Learning, as shown later.  $f_i^{\mathbf{obj}} \in \mathbb{R}^{d}$ is the vector representing the cluster centers in $F_i^{\mathbf{obj}}$, which can be viewed as a general representation for a set of collected $f_i^{\mathbf{pix}}$. It usually corresponds to a certain ground object or semantics. The object-level contrastive learning loss is computed as \cref{equ:object},
\vspace{-0.3em}
\begin{equation}
\begin{aligned}
\mathcal{L}_{\mathbf{obj}}(F_{i}, F_{i}^{\prime}) = \frac{1}{N_C T_i}\sum_s \sum_t \mathcal{L}_{CL}(f_i^{\mathbf{obj}}, f_i^{\mathbf{obj}\prime}).
\end{aligned}
\label{equ:object}
\end{equation}

$F_i^{\mathbf{img}} \in \mathbb{R}^{d}$ corresponds to the image-level feature, which is an average pooling result from $F_i^{\mathbf{pix}}$. Image-level contrastive learning loss is illustrated by \cref{equ:image},
\vspace{-0.3em}
\begin{equation}
\begin{aligned}
\mathcal{L}_{\mathbf{img}}(F_{i}, F_{i}^{\prime}) = \frac{1}{T_i}\sum_t \mathcal{L}_{CL}(F_i^{\mathbf{img}}, F_i^{\mathbf{img}\prime}).
\end{aligned}
\label{equ:image}
\end{equation}

The fine-grained contrastive learning loss $\mathcal{L}_{FGCL}$ is the sum of pixel-, object- and image-level contrastive learning losses as \cref{equ:fgcl}. Finally we form the Multi-Granularity Contrastive Learning loss $\mathcal{L}_{MGCL}$ in \cref{deqn_contras_modal}. The concept of multi-granularity is reflected in two aspects: space and modality. In terms of space, contrastive learning is performed at the pixel-, object-, and image-level, facilitating representation learning that encapsulates diverse spatial dimensions. Regarding modality, we conduct contrastive learning on the feature of each single modality, \ie, $F_i$, and the multi-modal feature after fusion, \ie, $F_{\mathbf{fus}}$,

\vspace{-0.3em}
\begin{equation}
\begin{aligned}
\mathcal{L}_{FGCL}(F_{i}, F_{i}^{\prime}) = \sum_{n \in \left\{\mathbf{pix}, \mathbf{obj}, \mathbf{img}\right\}} \mathcal{L}_{n}(F_{i}^{}, F_{i}^{\prime}),
\end{aligned}
\label{equ:fgcl}
\end{equation}
\vspace{-0.3em}
\begin{equation}
\begin{aligned}
\mathcal{L}_{MGCL} & = \sum_{i \in \left\{HR, Ms, SAR\right\}} \mathcal{L}_{FGCL}(F_{i}, F_{i}^{\prime}) \\
& + \mathcal{L}_{FGCL}(F_{\mathbf{fus}}, F_{\mathbf{fus}}^{\prime}).
\end{aligned}
\label{deqn_contras_modal}
\end{equation}

\noindent\textbf{Cross-Modal Alignment.} The heterogeneity of multi-modal data poses a challenge for effective multi-modal feature fusion. We address this issue by adopting multi-modal contrastive loss $\mathcal{L}_{MMCL}$ ~\cite{li2021align} to form the alignment loss $\mathcal{L}_{align}$, as shown in \cref{deqn_align},
\vspace{-0.3em}
\begin{equation}
\begin{aligned}
\mathcal{L}_{align} & =\sum_{i \neq j} \mathcal{L}_{MMCL}\left(F_{i}, F_{j}\right), \\
& i, j \in  \left\{HR, Ms, SAR\right\}.
\end{aligned}
\label{deqn_align}
\end{equation}
$\mathcal{L}_{MMCL}$ maximizes the similarity of cross-modal features from the same geo-location, while minimizing it otherwise. Cross-modal alignment is performed on the student branch.

\begin{table*}
    \begin{subtable}[t]{0.4\linewidth}
        \centering
        \setlength\tabcolsep{1.8pt}%
        \scriptsize
        \begin{tabular}{lccccc}
        \toprule
        \multirow{2}{*}{Model}     & \multirow{2}{*}{Publication}     & Dyna.-Pla. & iSAID & Potsdam & Dyna.-S2  \\ \cmidrule(lr){3-6}
                                   &                           & mIoU       & mIoU  & mF1     & mIoU      \\ \midrule
        GASSL \cite{ayush2021geography}                      & ICCV'21                   & 34.0/40.8     & 65.95 & 91.27   & 28.1/41.0 \\
        SeCo \cite{manas2021seasonal}                      & ICCV'21                   & -          & 57.20     & 89.03        & 29.4/39.8          \\
        SatMAE \cite{cong2022satmae}                    & NIPS'22                   & 32.8/39.9     & 62.97      & 90.63        & 30.1/38.7          \\
        RingMo$^{\dagger}$ \cite{sun2022ringmo}                    & TGRS'22                   & -          & 67.20      & 91.27        & -          \\
        RVSA \cite{wang2022advancing}                      & TGRS'22                   & 34.3/44.4     & 64.49      & -        & -          \\
        BFM$^{\dagger}$ \cite{cha2023billion}                        & Arxiv'23                 & -     & -      & 92.12        & -          \\
        TOV \cite{tao2023tov}                       & JSTARS'23                 & 32.1/37.8     & 66.24      & 92.03        & -          \\
        SSL4EO \cite{wang2022ssl4eo}                    & GRSM'23                   & 35.3/42.1     & 64.01      & 91.54        & 31.8/42.7          \\
        CMID \cite{muhtar2023cmid}                      & TGRS'23                   & 36.4/43.5  & 66.21      & 91.86        & -          \\
        CACo \cite{mall2023change}                      & CVPR'23                   & 35.4/42.7  & 64.32      & 91.35        & 30.2/42.5          \\
        SAMRS$^{\dagger}$ \cite{wang2023scaling}                 & NIPS'23                   & -     & 66.26      & 91.43        &  -         \\
        SatLas \cite{bastani2022satlas}                    & ICCV'23                   & 37.4/40.7  & 68.71      & 91.28        & 31.9/43.5          \\
        GFM \cite{mendieta2023gfm}   & ICCV'23         & 36.7/45.6        & 66.62                  & 91.85                 & -              \\
        Scale-MAE \cite{reed2022scale}                 & ICCV'23                   & 34.0/41.7     & 65.77      & 91.54        &  -         \\ \midrule
        \textbf{SkySense}                   & -                         & \textbf{39.7/46.5}           & \textbf{70.91}      & \textbf{93.99}        & \textbf{33.1/46.2}          \\ \bottomrule
        \end{tabular}
        \caption{Semantic segmentation results.}
        \label{tab:seg}
    \end{subtable}
    \hspace{0.3em}
    \begin{subtable}[t]{0.28\linewidth}
        \centering
        \setlength\tabcolsep{1.8pt}%
        \scriptsize
        \begin{tabular}{lccc}
        \toprule
        \multirow{3}{*}{Model}         & Horizontal & \multicolumn{2}{c}{Oriented} \\ \cmidrule(lr){2-2}\cmidrule(lr){3-4}
                                                             & DIOR       & DIOR-R        & FAIR1M       \\ \cmidrule(lr){2-4}
                                                             & mAP$_{50}$      & mAP        & mAP          \\ \midrule
        GASSL \cite{ayush2021geography}                                         & 67.40      & 65.65         & 48.15        \\
        SatMAE \cite{cong2022satmae}                                       & 70.89      & 65.66              & 46.55             \\
        RingMo$^{\dagger}$ \cite{sun2022ringmo}                    & 75.90                   & -      & 46.21                          \\
        RVSA \cite{wang2022advancing}                                         & 73.22      & 71.05              & 47.04             \\
        BFM$^{\dagger}$ \cite{cha2023billion}                                       & -          & 73.62              & -             \\
        TOV \cite{tao2023tov}                                        & 70.16      & 66.33              & 49.62             \\
        SSL4EO \cite{wang2022ssl4eo}                                       & 64.82      & 61.23              & 49.37             \\
        CMID \cite{muhtar2023cmid}                                         & 75.11      & 66.37              & 50.58             \\
        CACo \cite{mall2023change}                                         & 66.91      & 64.10              & 47.83             \\
        SatLas \cite{bastani2022satlas}                                       & 74.10      & 67.59              & 46.19             \\
        GFM \cite{mendieta2023gfm}      & 72.84        & 67.67                  & 49.69        \\
        Scale-MAE \cite{reed2022scale}                                    & 73.81      & 66.47              & 48.31             \\ \midrule
        \textbf{SkySense}                                           & \textbf{78.73}           & \textbf{74.27}              & \textbf{54.57}             \\ \bottomrule
        \end{tabular}
        \caption{Object detection results.}
        \label{tab:det}
    \end{subtable}
    \hspace{0.3em}
    \begin{subtable}[t]{0.28\linewidth}
        \centering
        \setlength\tabcolsep{1.8pt}%
        \scriptsize
        \begin{tabular}{lccc}
        \toprule
        \multirow{2}{*}{Model}  & LEVIR-CD & OSCD  & Dyna.-S2  \\ \cmidrule(lr){2-4}
                                                      & F1       & F1    & SCS       \\ \midrule
        GASSL \cite{ayush2021geography}                               & 78.19    & 46.26 & 13.6/16.7 \\
        SeCo \cite{manas2021seasonal}                                 & 90.14    & 47.67      &           13.9/16.0\\
        SatMAE \cite{cong2022satmae}                             & 87.65    & 52.76      &           14.8/16.2\\
        RingMo$^{\dagger}$ \cite{sun2022ringmo}                              & 91.86    & -      &           -\\
        RVSA \cite{wang2022advancing}                               & 90.86    & -      &           -\\
        SpectralGPT$^{\dagger}$ \cite{hong2023spectralgpt}
               & -        & 54.29      &           -\\
        MATTER$^{\dagger}$ \cite{akiva2022self}                               & -        & 59.37 &           -\\
        DINO-MC \cite{wanyan2023dino}                             & -        & 52.70 &           14.5/15.6\\
        SSL4EO \cite{wang2022ssl4eo}                               & 89.05    & 35.08      &           12.3/17.5\\
        CMID \cite{muhtar2023cmid}                                 & 91.72    & -      &           -\\
        CACo \cite{mall2023change}                                 & 81.04    & 52.11      &           15.3/15.8\\
        SatLas \cite{bastani2022satlas}                              & 90.62    & -      &           13.3/17.8\\
        GFM \cite{mendieta2023gfm}                                 & 91.73        & 59.82 &           -\\
        Scale-MAE \cite{reed2022scale}                           & 92.07    & -      &           -\\ \midrule
        \textbf{SkySense}                                   & \textbf{92.58}    & \textbf{60.06}      & \textbf{15.4/18.0}          \\ \bottomrule
        \end{tabular}
        \caption{Change detection results.}
        \label{tab:change}
    \end{subtable}
    \caption{Results of semantic segmentation, object detection and change detection. $\dagger$ means the code and weights are not released until November 11th, 2023, thus we report the metrics from the paper. - means the task is not supported or the value is unavailable in the paper.}
    \vspace{-1.2em}
\end{table*}

\begin{table}
\centering
\setlength\tabcolsep{4pt}%
\scriptsize
\begin{tabular}{lcccc}
\toprule
\multirow{3}{*}{Model} & \multicolumn{2}{c}{Single-label} & Multi-label & Temporal \\ \cmidrule(lr){2-3}\cmidrule(lr){4-4}\cmidrule(lr){5-5}
                       &  \begin{tabular}[c]{@{}c@{}}AID\\ \tiny(TR=20\%/50\%)\end{tabular}              & \begin{tabular}[c]{@{}c@{}}RESISC-45\\ \tiny(TR=10\%/20\%)\end{tabular}             & \begin{tabular}[c]{@{}c@{}}BEN-S2\\ \tiny(TR=10\%/100\%)\end{tabular}           & \begin{tabular}[c]{@{}c@{}}fMoW-S2\\ \tiny(TR=100\%)\end{tabular}       \\ \cmidrule(lr){2-5}
                       & OA                & OA                & mAP              & Top-1/5 Acc   \\ \midrule
GASSL \cite{ayush2021geography}                  & 93.55/95.92       & 90.86/93.06       & 79.24/87.40            & 50.69/77.99   \\
SeCo \cite{manas2021seasonal}                   & 93.47/95.99       & 89.64/92.91       & 82.62/87.81            & 51.65/77.40   \\
SatMAE \cite{cong2022satmae}                & 95.02/96.94       & 91.72/94.10                  & 86.18/89.50                 & 63.84/-              \\
RingMo$^{\dagger}$ \cite{sun2022ringmo}                & 96.90/98.34       & 94.25/95.67                  & -                 & -              \\
RVSA \cite{wang2022advancing}                  & 97.03/98.50       & 93.93/95.69                  & -                 & -              \\
DINO-MC \cite{wanyan2023dino}               & -                  & -                  & 84.20/88.75            & 60.16/83.49   \\
TOV \cite{tao2023tov}                   & 95.16/97.09       & 90.97/93.79                  & -                 & -              \\
SSL4EO \cite{wang2022ssl4eo}                & 91.06/94.74       & 87.60/91.27                  & 87.10/91.80                 & 51.70/76.77              \\
CMID \cite{muhtar2023cmid}                  & 96.11/97.79       & 94.05/95.53                  & -                 & -              \\
CACo \cite{mall2023change}                  & 90.88/95.05       & 88.28/91.94                  & 81.30/87.00                 & 50.72/76.31              \\
CROMA$^{\dagger}$ \cite{fuller2023croma}  & -                 & -                            & 88.29/-           & 63.59/-         \\ 
SatLas \cite{bastani2022satlas}                & 94.96/97.38       & 92.16/94.70                  & 82.80/88.37                 & 57.95/79.00              \\
GFM \cite{mendieta2023gfm}                  & 95.47/97.09        & 92.73/94.64                  & 86.30/-                 & -              \\
Scale-MAE \cite{reed2022scale}             & 96.44/97.58       & 92.63/95.04                  & -                 & -              \\ \midrule
\textbf{SkySense}               & \textbf{97.68/98.60}       & \textbf{94.85/96.32}                  & \textbf{88.67/92.09}                 & \textbf{64.38/87.27}              \\ \bottomrule
\end{tabular}
\caption{Scene classification results.}
\vspace{-1.2em}
\label{tab:cla}
\end{table}

\noindent\textbf{Unsupervised Geo-Context Prototype Learning.} 
Different regions characterize distinct geographic landscapes and seasonal dynamics~\cite{huang2022toward,hu2020unsupervised} due to disparities in topography and climate. Prior arts have shown that the enlarged context can benefit the RSI interpretations~\cite{liu2023seeing,chen2019collaborative,guo2022isdnet,huang2022toward}. In this work, $F_{\mathbf{fus}}^{\text{mm}}$ captures rich spatiotemporal clues for a small area. By clustering on numerous $F_{\mathbf{fus}}^{\text{mm}}$, higher-level regional semantics are obtained as implicit geo-knowledge for a vast geo-spatial scope (see \cref{fig:gcpl}). Thus, we propose Geo-Context Prototype Learning to unsupervisedly extract regional geo-context from $F_{\mathbf{fus}}^{\text{mm}}$ during pre-training.

We divide the globe into $N_{R}$ regions and initialize a region-specific prototype set $\mathcal{P} \in \mathbb{R}^{N_{R} \times N_{p} \times d}$. Each prototype is learned from $F_{\mathbf{fus}}^{\text{mm}}$. We leverage the geo-location of the RSI to retrieve the regional subset $\mathcal{P}_r \in \mathbb{R}^{N_{p} \times d}$  from $\mathcal{P}$. Then, we calculate the cosine similarity matrix $\mathbf{M} \in \mathbb{R}^{N_{S} \times N_{p}}$ between $F_{\mathbf{fus}}^{\text{mm}}$ and $\mathcal{P}_r$ as in \cref{deqn_cosm},
\vspace{-0.4em}
\begin{align}
\label{deqn_cosm}
\mathbf{M} = \frac{F_{\mathbf{fus}}^{\text{mm}} \cdot \mathcal{P}_r^{\text{T}}}{\Vert F_{\mathbf{fus}}^{\text{mm}} \Vert \Vert \mathcal{P}_r \Vert},
\end{align}

We utilize the Sinkhorn-Knopp algorithm \cite{caron2020unsupervised} on $\mathbf{M}$ to find the optimal assignment matrix $\mathbf{S} \in \mathbb{R}^{N_{S} \times N_{p}}$ between $F_{\mathbf{fus}}^{\text{mm}}$ and the prototypes. This algorithm introduces the uniform distribution constraint to avoid trivial solution while achieving maximal similarity possible. We then use $\mathbf{S}$ to generate an update value for current sample's corresponding $\mathcal{P}_r$, denoted as $\overline{\mathcal{P}_r}$, as shown in \cref{deqn_ex2},

\vspace{-0.4em}
\begin{align}
\label{deqn_ex2}
\overline{\mathcal{P}_r} = \mathbf{S}^{\text{T}}F_{\mathbf{fus}}^{\text{mm}}.
\end{align}
Afterwards, we update $\mathcal{P}_r$ through EMA~\cite{he2020momentum} as in \cref{deqn_ex3}, where $m \in [0, 1)$ is a momentum coefficient,
\vspace{-0.4em}
\begin{align}
\label{deqn_ex3}
\mathcal{P}_{r}\leftarrow m \mathcal{P}_r + (1-m) \overline{\mathcal{P}_r}.
\end{align}

Each $\mathcal{P}_{r}$ is updated during pre-training and used as the fixed geo-context for downstream tasks. Geo-Context Prototype Learning is only conducted on the student branch. It extracts generalized region-aware representations from numerous RSI within a consistent region, offering a complementary clue to enhance the feature of a single RSI.

As Geo-Context Prototype Learning is incorporated without an explicit loss term, our pre-training objective is shown in \cref{deqn_ex5}, where $\alpha$ and $\beta$ are trade-off weights,
\vspace{-0.3em}
\begin{align}
\label{deqn_ex5}
\mathcal{L} = \alpha\mathcal{L}_{MGCL} + \beta\mathcal{L}_{align}.
\end{align}
\section{Experiments}
\label{sec4}
\cref{fig:fig1} demonstrates SkySense's superior performance in all test scenarios. We conduct experiments on 16 datasets, covering different modalities and tasks, to ensure a comprehensive assessment. The right side of \cref{fig:fig3} shows how to apply SkySense to different tasks. Each pre-trained module is designed to allow for combined or individual use, with the flexibility to be either frozen or fine-tuned as needed. More details are included in the supplementary materials.

\subsection{Pre-training Implementation}\label{pretraindetails}

The model is pre-trained with a batch size of 240 samples, distributed over 80 A100-80GB GPUs. For HSROIs, we apply data augmentations including multi-crop~\cite{caron2020unsupervised}, Gaussian blur, solarization~\cite{grill2020bootstrap}, \etc. As for TMsI and TSARI, we randomly select a fixed-sized sequence from the original one and perform random disturbances on the RSI acquisition date. We employ the huge version of the Swin Transformer (Swin-H)~\cite{liu2021swin} as the spatial encoder of HSROIs, for its design efficiency in minimizing computational costs for high-resolution imagery~\cite{zhang2021multi}. RSI from TMsI or TSARI is processed with corresponding ViT-L~\cite{dosovitskiy2020image}. For Geo-Context Prototype Learning, we divide the globe into 4096 regions, each containing 100 prototypes.

\subsection{Performance on Single-Modal Tasks}
We evaluate SkySense on 4 representative single-modal tasks. All experiments are conducted using consistent fine-tuning settings for fairness. The supplementary materials include implementation details, visualization results and additional experiments on frozen backbone tuning.

\noindent\textbf{Semantic Segmentation.} We adopt Dyna.-Pla.~\cite{toker2022dynamicearthnet}, iSAID~\cite{waqas2019isaid}, Potsdam~\cite{sherrah2016fully} and Dyna.-S2~\cite{toker2022dynamicearthnet} for the segmentation experiment. They are chosen considering factors such as spatial resolution, spectrum and category type. 
UperNet~\cite{xiao2018unified} serves as the segmentation head. For Dyna.-Pla. and Dyna.-S2 datasets, we report mIoU results on official validation and test sets. For iSAID and Potsdam, we follow the settings of \cite{sun2022ringmo}. As depicted in \cref{tab:seg}, SkySense has achieved the SOTA performance on all four segmentation datasets. On average, it surpasses the previous SOTA by an impressive improvement of 1.86\%.

\noindent\textbf{Horizontal \& Oriented Object Detection.}
We employ the widely recognized DIOR~\cite{li2020object} dataset for Horizontal Object Detection and its enhanced version DIOR-R~\cite{cheng2022anchor}, along with FAIR1M~\cite{sun2022fair1m}, for Oriented Object Detection. All datasets consist of optical RGB images. Faster RCNN~\cite{ren2015faster} and Oriented RCNN~\cite{li2022oriented} are used for the experiment, following the setup of~\cite{sun2022ringmo, wang2022advancing}. SkySense excels on all three datasets (\cref{tab:det}). Notably, we surpass the second best CMID by 3.99\% mAP and have achieved the best performance on the FAIR1M v2.0\footnote{https://www.gaofen-challenge.com/benchmark \textit{(2023.11.17)}} leaderboard. More importantly, our results are accomplished without using any sophisticated Oriented Detection designs~\cite{hou2022shape,li2022oriented,yu2023phase}.

\noindent\textbf{Change Detection.} We assess SkySense's Change Detection performance on LEVIR-CD~\cite{chen2020spatial}, OSCD~\cite{daudt2018urban}, and Dyna.-S2~\cite{toker2022dynamicearthnet} datasets. For LEVIR-CD and OSCD, we follow the frameworks of \cite{sun2022ringmo, wanyan2023dino} and report the F1 metric. For Dyna.-S2, we utilize the UperNet head since the reported semantic change segmentation (SCS) score is calculated from the segmentation results~\cite{toker2022dynamicearthnet}. Both results from the validation and test sets of Dyna.-S2 are presented. The remarkable generalization ability of SkySense is evident in the consistent improvements shown in \cref{tab:change}. Unlike CACo~\cite{mall2023change}, which shows proficiency mainly on the Dyna.-S2 validation set, our model excels across all datasets.

\noindent\textbf{Scene Classification.} We utilize four scene classification datasets: AID~\cite{xia2017aid} and RESISC-45~\cite{cheng2017remote} with static RGB images, BEN-S2~\cite{sumbul2020bigearthnet} with static multispectral images, and fMoW-S2~\cite{cong2022satmae} with temporal multispectral images. The training ratio (TR) follows~\cite{sun2022ringmo,manas2021seasonal,cong2022satmae}. We use a linear classifier head for experiment. For AID and RESISC-45, we report Overall Accuracy (OA), for BEN-S2 we report mAP, and for fMoW-S2 we report both Top-1 and Top-5 Accuracy. SkySense overally outperforms competitive baselines and achieves the best results on all datasets (\cref{tab:cla}). Additionally, with limited labeled data on the AID dataset, SkySense consistently outperforms CMID, Scale-MAE, and random initialization, with a 4.17\% higher OA than the second best Scale-MAE using only 1\% training data (\cref{fig:cloud}\textcolor{red}{a}). These results highlight the robustness and generalization ability of SkySense's pre-trained features.

\begin{figure}
	\centering
		\includegraphics[scale=.3]{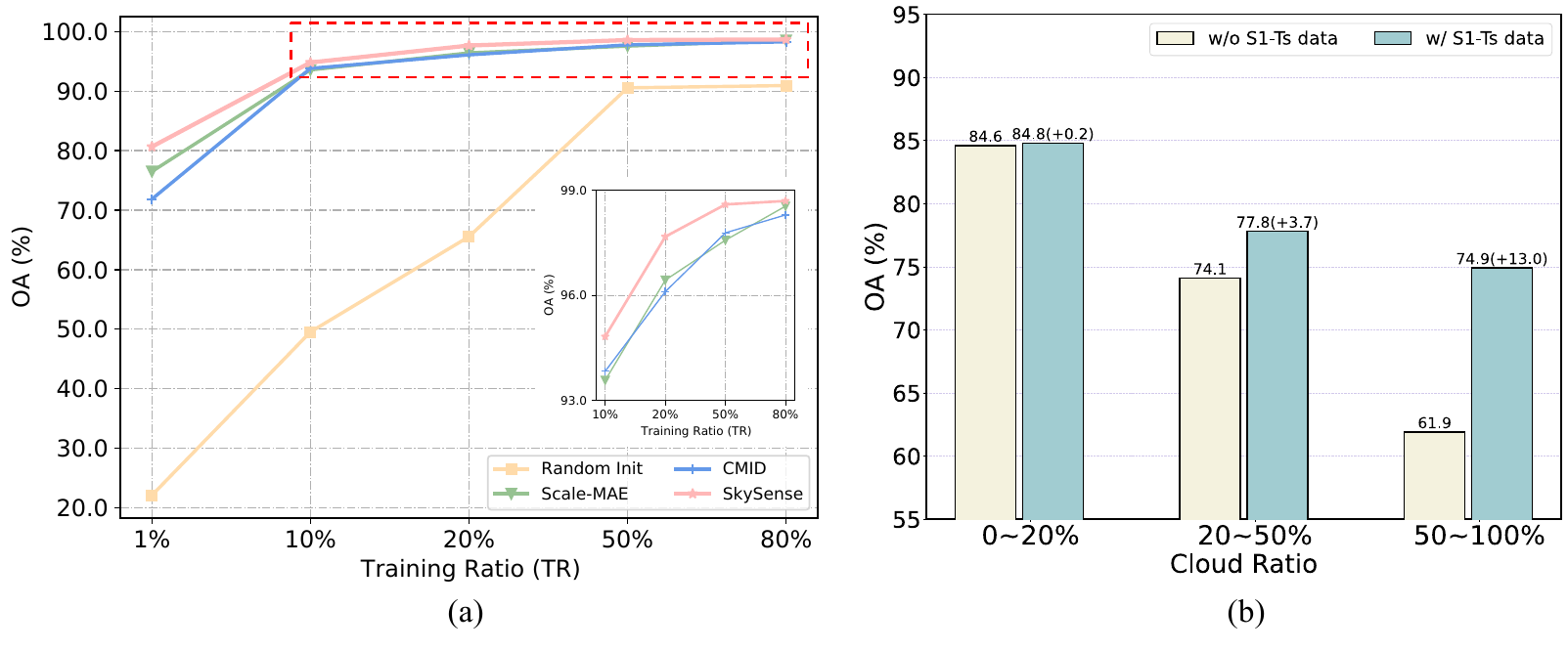}
        \vspace{-0.8em}
	    \caption{(a) Experiment on fine-tuning using different percentages of training data on the AID dataset. (b) The impact of S1-Ts data under varying cloud coverage conditions.}
        \vspace{-0.5em}
	\label{fig:cloud}
\end{figure}

\begin{table}
\centering
\setlength\tabcolsep{2.5pt}%
\scriptsize
\begin{tabular}{llcc}
\toprule
\textbf{\begin{tabular}[c]{@{}l@{}}Task \& Dataset\end{tabular}}                                             & \textbf{Data Source \& Geo-Context}   & \textbf{\begin{tabular}[c]{@{}c@{}}Previous \\ SOTA\end{tabular}} & \textbf{SkySense}  \\ \midrule
\multirow{6}{*}{\begin{tabular}[c]{@{}l@{}}(a) Multi-Modal Seg:\\ Dyna.-MM\end{tabular}}                        & \texttt{(i)} Planet.             & 45.6 \cite{mendieta2023gfm}                                                                 &  46.5 \textbf{\textcolor[rgb]{0.25, 0.5, 0.75}{$\uparrow$ 0.9}}              \\

              & \texttt{(ii)} Planet. with GCP & - & 47.0 \\
                                                                                                                & \texttt{(iii)} S2                  & 43.5 \cite{bastani2022satlas}                                                                  & 46.2 \textbf{\textcolor[rgb]{0.25, 0.5, 0.75}{$\uparrow$ 2.7}}              \\
                                                                                                                & \texttt{(iv)} Planet. + S2   & -                                                                  &                  47.3          \\
                                                                                                                & \texttt{(v)} Planet. + S2 + S1   & -                                                                &                   47.7           \\
                                                                                                                & \texttt{(vi)} Planet. + S2 + S1 with GCP & - & 48.2 \\ \midrule
\multirow{5}{*}{\begin{tabular}[c]{@{}l@{}}(b) Multi-Modal Seg: \\ PASTIS-MM\end{tabular}} &  \texttt{(i)} S2-static  & - & 73.5 \\

 & \texttt{(ii)} S2-Ts               & 83.4 \cite{tarasiou2023vits}                                                                  &  84.6 \textbf{\textcolor[rgb]{0.25, 0.5, 0.75}{$\uparrow$ 1.2}}              \\

                                                                                                                & \texttt{(iii)} S2-Ts + S1-Ts       & 84.2 \cite{garnot2022multi}                                                                  &   84.8   \textbf{\textcolor[rgb]{0.25, 0.5, 0.75}{$\uparrow$ 0.6}}            \\
                                                                                                                & \texttt{(iv)} S2-Ts + GEP         & -                                                                   & 85.8                 \\
                                                                                                                & \texttt{(v)} S2-Ts + S1-Ts + GEP & -                                                                   & 85.9                \\ \midrule
\multirow{2}{*}{\begin{tabular}[c]{@{}l@{}}(c) Multi-Modal Cls: \\ BEN-MM\end{tabular}} &  \texttt{(i)} S1  & 83.70~\cite{wang2023decur} & 86.25 \textbf{\textcolor[rgb]{0.25, 0.5, 0.75}{$\uparrow$ 2.55}} \\
&  \texttt{(ii)} S2 + S1  & 89.70~\cite{wang2023decur} & 92.21\textbf{\textcolor[rgb]{0.25, 0.5, 0.75}{$\uparrow$ 2.51}} \\ \bottomrule
\end{tabular}
\caption{Fine-tuning results on multi-modal tasks.}
\vspace{-1em}
\label{tab6}
\end{table}

\subsection{Performance on Multi-Modal Tasks}\label{MMDT}
\noindent\textbf{Multi-Modal Segmentation: Time-insensitive Land Cover Mapping.} We employ the Dyna.-MM dataset~\cite{toker2022dynamicearthnet} for fine-tuning and report mIoU on the official test set. The dataset comprises HSROIs from PlanetFusion (Planet.), multispectral imagery from Sentinel-2 (S2), and SAR imagery from Sentinel-1 (S1). We use a simple UperNet head. As shown in \cref{tab6}\textcolor{red}{a}, SkySense achieves the best results in single-modal scenarios \texttt{(i)} and \texttt{(iii)}, clearly outperforming the previous SOTA by roughly 1\% mIoU. Moreover, combining all three modalities as \texttt{(v)} further improves the mIoU by 1.2\% compared to \texttt{(i)}. Notably, without bells and whistles, SkySense ranks No.1 on the challenging DynamicEarthNet leaderboard\footnote{https://codalab.lisn.upsaclay.fr/competitions/2882\#results\textit{(2023.11.17)}}. 

\noindent\textbf{Multi-Modal Segmentation: Time-sensitive Crop Mapping.} We evaluate SkySense's fine-tuning result on the PASTIS-MM dataset, an enhanced version of PASTIS-R~\cite{garnot2022multi}. PASTIS-MM includes HSROIs from Google Earth Pro (GEP), TMsI from Sentinel-2 (S2-Ts), and TSARI from Sentinel-1 (S1-Ts). We use a naive FCN head~\cite{long2015fully} and report the OA from the official five-fold validation on PASTIS-MM dataset. In \cref{tab6}\textcolor{red}{b}, comparing S2-Ts and static multispectral data (S2-static), we observe a significant 11.1\% OA increase, highlighting the importance of incorporating temporal clue for crop mapping.

Furthermore, both \texttt{(ii)} and \texttt{(iii)} exceed the performance of the previous SOTA, affirming the superior capabilities of SkySense. When more modalities are added as \texttt{(ii)}, \texttt{(iv)}, and \texttt{(v)}, the OA increases accordingly. However, integrating Sentinel-1 data yields no substantial improvement, presumably because of the cloud-free imagery from PASTIS-MM dataset. To further investigate, we compare OA using S2-Ts data at different cloud ratios with and without Sentinel-1. \cref{fig:cloud}\textcolor{red}{b} illustrates that the performance difference between utilizing and foregoing Sentinel-1 data becomes more pronounced with an increasing cloud ratio. Specifically, when the cloud ratio exceeds 50\%, the result of using Sentinel-1 outperforms its counterpart by 13\%. This highlights the importance of SAR data in situations with cloud coverage and rainfall.

\noindent\textbf{Multi-Modal Scene Classification.} We utilize the BEN-MM~\cite{Sumbul2021bigearthnet} dataset for evaluating the multi-modal scene classification task. This dataset includes both Sentinel-1 (S1) and Sentinel-2 (S2) imagery. The evaluation protocol from DeCUR~\cite{wang2023decur} is followed, and we report the mAP metric of fine-tuning with $100\%$ training data. In \cref{tab6}\textcolor{red}{c}, both \texttt{(i)} and \texttt{(ii)} significantly outperform the previous SOTA by more than $2.5\%$ mAP. Furthermore, the inclusion of Sentinel-2 imagery greatly enhances performance compared to using Sentinel-1 imagery alone.

All these results show a notable gain for the tasks using multi-modal data, affirming the necessity of SkySense's multi-modal pre-training from one perspective.

\section{Discussions \& Ablation Studies}
\noindent\textbf{Multi-modal Pre-training Effectiveness.} In addition to confirming the effectiveness of using multi-modal data in downstream tasks, we investigate the impact of multi-modal pre-training on single-modal tasks, compared with pre-training on fewer modalities. We conduct experiments on iSAID for static HSROI segmentation and fMoW-S2 for temporal multispectral classification. Two versions of the pre-trained model are tested: a simple version pre-trained only with optical imagery (HSROIs, TMsI), and SkySense, which includes HSROIs, TMsI, and TSARI for pre-training. The rest of the settings remain consistent. The results in \cref{modalexp} show that SkySense consistently outperforms the simple version, suggesting that the introduction of SAR data benefits representation learning of other modalities. This may attribute to the implicit clue brought by SAR data through Cross-Modal Alignment. It provides another perspective on the necessity of SkySense's multi-modal pre-training.

\begin{table}
    \centering
    \begin{subtable}[t]{0.495\linewidth}
        \setlength\tabcolsep{2pt}%
        \centering
        \scriptsize
        \begin{tabular}{lcc}
        \toprule
        \textbf{\multirow{2}{*}{Pre-training}}    & \textbf{iSAID}  & \textbf{fMoW-S2} \\\cmidrule(lr){2-2}\cmidrule(lr){3-3}     & mIoU       & Top-5 Acc     \\\midrule
        Simple Ver.          & 68.98         & 85.69      \\
        SkySense       & 70.91 \textbf{\textcolor[rgb]{0.25, 0.5, 0.75}{$\uparrow$ 1.93}}      & 87.27 \textbf{\textcolor[rgb]{0.25, 0.5, 0.75}{$\uparrow$ 1.58}}      \\ \bottomrule
        \end{tabular}
        \caption{}
        \label{modalexp}
    \end{subtable}
    \begin{subtable}[t]{0.495\linewidth}
        \centering
        \setlength\tabcolsep{6pt}%
        \scriptsize
        \begin{tabular}{lc}
        \toprule
        \textbf{\multirow{2}{*}{Pre-training}} & \textbf{Dyna.-MM} \\\cmidrule(lr){2-2} & mIoU \\ \midrule
        Baseline            & 42.2                     \\
        + MGCL                      & 44.4 \footnotesize{\textcolor[rgb]{0.25, 0.5, 0.75}{$\uparrow$ 2.2}}                      \\
        + MM & 47.0 \footnotesize{\textcolor[rgb]{0.25, 0.5, 0.75}{$\uparrow$ 2.6}}  \\ 
        + CMA & 47.7 \footnotesize{\textcolor[rgb]{0.25, 0.5, 0.75}{$\uparrow$ 0.7}}  \\ 
        + GCPL & 48.2 \footnotesize{\textcolor[rgb]{0.25, 0.5, 0.75}{$\uparrow$ 0.5}}  \\ \bottomrule
        \end{tabular}
        \caption{}
        \label{tab8}
    \end{subtable}
    \caption{(a) Discussion on multi-modal pre-training effectiveness. (b) Ablation study on the pre-training design.}
    \vspace{-1.2em}
\end{table}

\noindent\textbf{What does Geo-Context Prototype (GCP) Learn?} We utilize the Dyna.-MM dataset for experiment as it contains diverse geo-locations worldwide. For the segmentation task in \cref{tab6}\textcolor{red}{a}, adding GCP in downstream tasks leads to a further gain of 0.5\% mIoU compared to the strong multi-modal baseline \texttt{(v)}. Moreover, a comparison between \texttt{(i)} and \texttt{(ii)} shows a 0.5\% mIoU improvement using GCP for the single-modal task. It demonstrates GCP's consistent performance gain in single- and multi-modal scenarios.

In \cref{fig:gcpl}, we visualize the learned prototypes on the Map by calculating the pre-trained feature of each pixel and assigning the most similar prototype to it. A comparison with the ESRI LandCover Map~\cite{karra2021global} reveals GCP's promising results in segmenting different areas. Moreover, GCP exhibit fine-grained advantage, as shown in the middle of \cref{fig:gcpl}. The prototypes learned from unsupervised clustering segment cropland within the town, which is overlooked by the LandCover Map. Notably, the visualization shares the same spatial resolution with ESRI LandCover Map. 

\begin{figure}
	\centering
		\includegraphics[scale=.115]{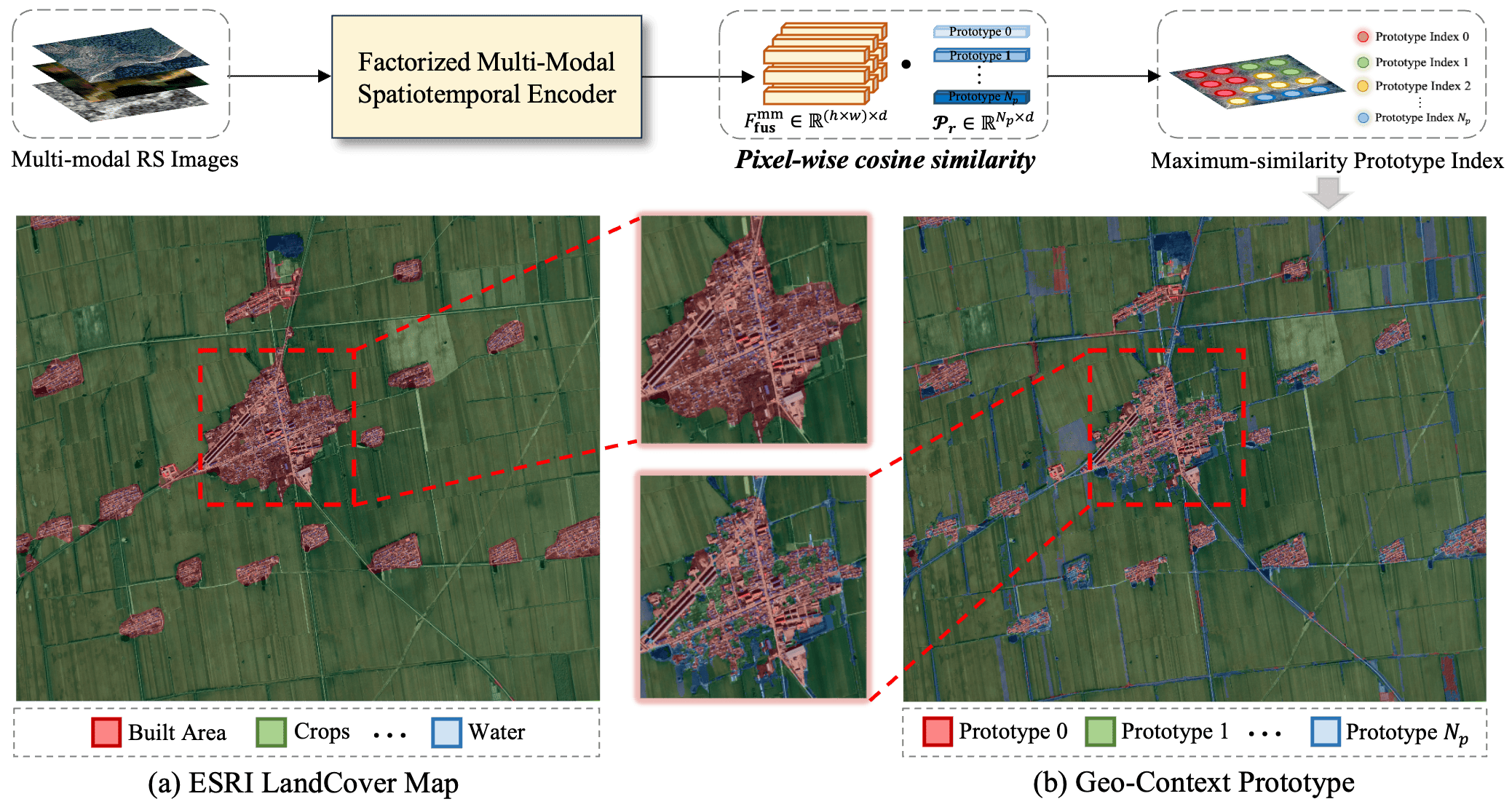}
        \vspace{-1.6em}
	    \caption{Comparison between (a) ESRI LandCover Map and (b) Geo-Context Prototype. The visualization process of Geo-Context Prototype is illustrated in the upper part of this figure.}
    \vspace{-1em}
	\label{fig:gcpl}
\end{figure}

\noindent\textbf{Design of Pre-training.} \cref{tab8} presents the ablation study to assess our pre-training design, namely Multi-Granularity Contrastive Learning (MGCL), Multi-Modal (MM) integration, Cross-Modal Alignment (CMA) and Geo-Context Prototype Learning (GCPL). We utilize the Dyna.-MM dataset for the experiment and report mIoU metric on the official test set. 

Initially, we utilize a single-modal version, using $g_{HR}$ spatial encoder and HSROIs for pre-training. The training settings are kept the same as described in \cref{pretraindetails}. The results show that MGCL leads to a notable improvement compared to a simple baseline~\cite{caron2021emerging}. Then we integrate further multi-modal data (\ie, TMsI and TSARI) into the pre-training and downstream evaluation. This effectively improves the performance on the test set to 47.0\% mIoU, validating the necessity of multi-modal pre-training. 

CMA is another necessary design for SkySense's pre-training, which explicitly pulls features from different modalities together, encouraging cross-modal interactions. The results show that the incorporation of modal alignment leads to 0.7\% mIoU improvement. Finally, we introduce GCPL, which learns complementary regional context clue to facilitate downstream tasks and further pushes the very strong performance to 48.2\% mIoU.
\section{Conclusion \& Future Work}
In this paper, we present SkySense, a large-scale MM-RSFM for interpretation of EO imagery. SkySense allows using its modules flexibly to accommodate different scenarios and consistently outperforms other models on a variety of tasks, showcasing its exceptional generalization ability and strong performance. We hope SkySense will inspire further research on MM-RSFM and its release may contribute to sustainable innovations thriving in the Remote Sensing community. As part of our future work, we plan to incorporate the language modality, thereby extending SkySense's applications to more EO tasks.

\section{Acknowledgment}
This work was in part supported by the National Natural Science Foundation of China under Grants 42030102 and 42371321.
{
    \small

    \bibliographystyle{ieeenat_fullname}
    \bibliography{main}
}

\clearpage
\maketitlesupplementary
\setcounter{section}{0}
\renewcommand\thesection{\Alph {section}}
\renewcommand\thefigure{\Alph{section}\arabic{figure}} 
\renewcommand\thetable{\Alph{section}\arabic{table}}

\section{Overview}
We provide the following materials to supplement our paper and divide them into 12 sections.
\begin{itemize}

\item To demonstrate the exceptional generalization capability of SkySense's pre-trained features, we assess the model's performance on multiple downstream datasets using frozen backbone features, as detailed in \cref{sup:sec:frozen}. Additionally, we present results from datasets acquired using different satellite sensors compared to our pre-training data in the same section, \cref{sup:sec:frozen}.

\item In \cref{sup:sec:convergence}, we execute a comparative analysis to assess the convergence rates and verify the efficacy of the SkySense in leveraging the features acquired through pre-training.

\item In \cref{sup:sec:swinl}, we conduct a comparison on the SkySense with different numbers of parameters to explore the influence of model scale.

\item We compare our model with randomly initialized counterpart and the representative Vision Foundation Model, DINOv2, as described in \cref{sup:sec:dino}.

\item We perform comparative experiments to establish the superior performance of the fundamental unit of SkySense, Factorized Spatiotemporal Encoder in \cref{sup:sec:factorized}.

\item In \cref{sup:sec:alignment}, we illustrate the feature visualization results to further analyze the effectiveness of Cross-Modal Alignment, providing the evidence for improved multi-modal feature fusion. Moreover, we provide the experiment result on pre-training with MAE to complement our ablation study, validating our design choice for NOT including MAE.

\item \cref{sup:sec:visual} presents qualitative results showcasing a range of visual comparisons, thereby providing further evidence to support the superiority of our method.

\item In \cref{sup:sec:dataset}, we introduce the details of our pre-training dataset construction and showcase illustrative remote sensing tile examples for better understanding.

\item In \cref{sup:sec:implementation}, we elaborate on the details of our SkySense pre-training implementation and its pre-training cost.

\item The exposition of the downstream datasets and the corresponding fine-tuning implementation settings is elucidated in \cref{sup:sec:dowmstream}.

\item Finally, we illustrate the procedure on how to use our SkySense pre-trained weights, and provide a simple scene classification example in \cref{sup:sec:example}.

\end{itemize}

\section{Experimental results of the frozen backbone tuning and various satellite sensors}
\label{sup:sec:frozen}
To validate the superiority of the features learned through our pre-training, we carry out experiments using \textbf{frozen} backbone features. Specifically, we tune task-specific heads while keeping the backbone parameters fixed for three downstream tasks: scene classification on the AID dataset \cite{xia2017aid}, object detection on the DIOR dataset \cite{li2020object}, and semantic segmentation on the iSAID dataset \cite{waqas2019isaid}.
The heads chosen for these tasks are the same as our paper's main experiment (the configuration outlined in \cref{sup:sec:dowmstream}).
We compare SkySense with four recent works, \ie~, CMID \cite{muhtar2023cmid}, SatLas \cite{bastani2022satlas}, GFM \cite{mendieta2023gfm}, and Scale-MAE \cite{reed2022scale}, the results are shown in \cref{tab:frozen_backbone}.
On the AID dataset, SkySense shows impressive advantages over other methods. Especially on the setting of less training data (\ie, 20\% data for training and the rest for testing), SkySense achieves notable improvements in overall accuracy (OA) compared to other approaches. Specifically, compared to SatLas, SkySense achieves a 28.09\% increase in OA. Similarly, when compared to Scale-MAE, SkySense surpasses it by 17.64\%.
Furthermore, SkySense exhibits a significant OA improvement of 14.65\% over GFM and a noteworthy enhancement of 6.27\% over CMID.
The results obtained from the DIOR and iSAID datasets further confirm that SkySense outperforms other methods.
Remarkably, our method shows a substantial average improvement of 9.69\% over other methods when evaluated on the iSAID dataset. 
The aforementioned experimental results provide compelling evidence that our SkySense achieves superior performance with the fixed backbone. This finding reinforces the notion that our model has successfully acquired more generalized features through pre-training.

To further validate SkySense's generalizability across multiple satellite sensors other than the data used for pre-training, we conduct experiments on three additional datasets from Gaofen and Landsat satellites, as shown in \cref{sensor}. SkySense greatly outperforms the other RSFMs.
\begin{table*}[htbp]
\centering
\begin{tabular}{lcccc}
\toprule
\multirow{2}{*}{Model}  & \multirow{2}{*}{Publication}  & AID               & DIOR    & iSAID \\ \cmidrule(lr){3-5}
              &         & OA (TR=20\% / 50\%) & $\mathrm{mAP}_{50}$ & mIoU  \\ \midrule
CMID \cite{muhtar2023cmid}   &   TGRS'23       & 87.80/90.92       & 66.08   & 59.40  \\
SatLas \cite{bastani2022satlas}  &  ICCV'23    & 65.98/75.02       & 60.46   & 56.03 \\
GFM \cite{mendieta2023gfm}  &  ICCV'23    & 79.42/87.37       & 67.34   & 60.86 \\
Scale-MAE \cite{reed2022scale}  &  ICCV'23    & 76.43/87.81       & 70.55   & 46.53 \\
\midrule
\textbf{SkySense}        &  -     & \textbf{94.07/95.85}       & \textbf{72.54}   & \textbf{65.40} \\ \bottomrule
\end{tabular}
\caption{Frozen backbone tuning results for downstream tasks.}
\label{tab:frozen_backbone}
\end{table*}

\begin{table*}[htbp]
\centering
\begin{tabular}{cccc}
\toprule
Dataset & Sensor & Previous Best RSFM & SkySense\\
\midrule
Five-Billion-Pixels\cite{FBP2023} & Gaofen-2 & 69.31 (Scale-MAE) & \textbf{74.46}\\
SPARCS\cite{sparcs} & Landsat-8 & 66.84 (GFM) & \textbf{72.88}\\
AIR-PolSAR-Seg\cite{airpolsar} & Gaofen-3 (SAR)  & 53.90 (CROMA) & \textbf{56.04}\\
\bottomrule
\end{tabular}
\caption{Results on datasets built from various sensors. (mIoU)}
\label{sensor}
\end{table*}

\section{Comparison of convergence rates}
\label{sup:sec:convergence}

\begin{figure*}
  \centering
  \begin{subfigure}{0.33\linewidth}
    \centering
    \includegraphics[width=\linewidth]{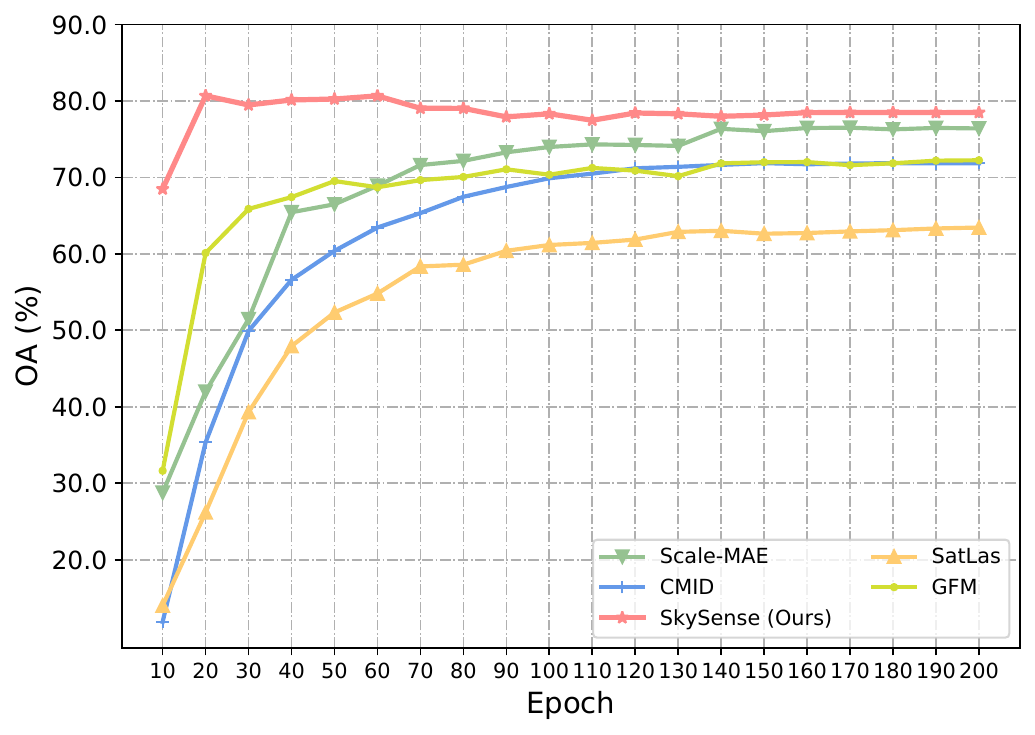}
    \caption{}
    \label{fig:conver_aid}
  \end{subfigure}
  \hfill
  \begin{subfigure}{0.33\linewidth}
    \centering
    \includegraphics[width=\linewidth]{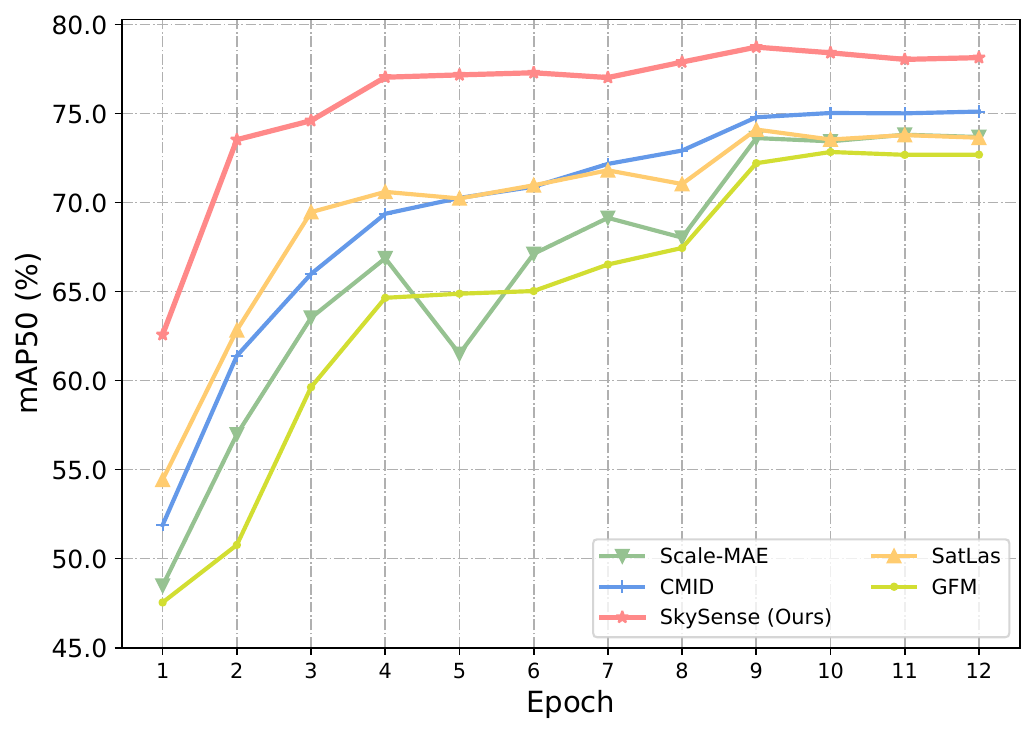}
    \caption{}
    \label{fig:conver_dior}
  \end{subfigure}
  \hfill
  \begin{subfigure}{0.33\linewidth}
    \centering
    \includegraphics[width=\linewidth]{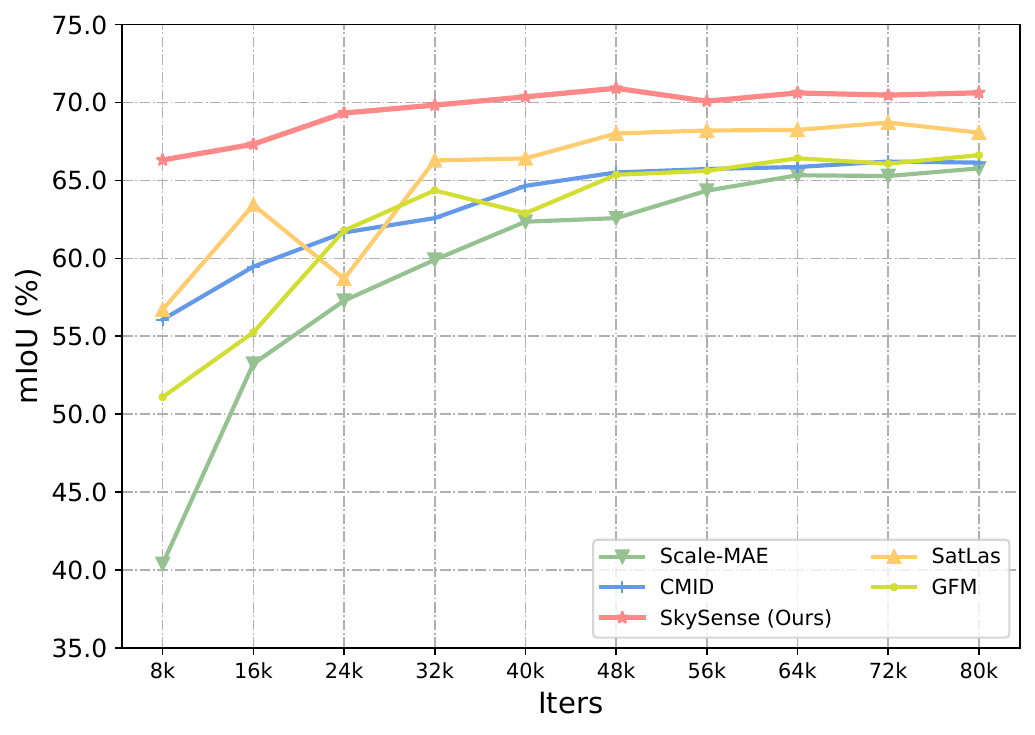}
    \caption{}
    \label{fig:conver_isad}
  \end{subfigure}
  \caption{Convergence curves of different methods on downstream tasks: (a) scene classification on the AID dataset (TR=1\%), (b) object detection on the DIOR dataset, and (c) semantic segmentation on the iSAID dataset.}
  \label{fig:conver_rates}
\end{figure*}

The convergence rate in downstream tasks serves as a pivotal metric in comprehensively evaluating a foundational model. In essence, a well-learned and robust feature representation during the pre-training phase facilitates swift convergence and enhances the model's overall efficacy in downstream tasks \cite{cha2023billion}.

To facilitate a comprehensive comparison between the SkySense and other Remote Sensing Foundation Models (RSFMs), we conduct a comparison of their convergence rates in different tasks.
Specifically, we perform comparative experiments between SkySense and recent approaches, such as Scale-MAE, Satlas, CMID and GFM, in three downstream tasks: scene classification, object detection, and semantic segmentation.
The convergence curves of the these models are depicted in \cref{fig:conver_rates}.
The results indicate that under the same experimental settings, SkySense exhibits the fastest convergence rate in all three downstream tasks. Particularly noteworthy is its performance in the scene classification task on the AID dataset (TR=1\%), where SkySense achieves desirable results with only 20 epochs. Other models require at least 140 epochs to converge to a stable but lower result. This remarkable advantage in convergence rates serves as the evidence that SkySense has successfully captured and encoded valuable information in its feature representations through effective pre-training, enabling it to rapidly adapt and generalize to downstream tasks.

\section{Experimental results of SkySense with fewer parameters}
\label{sup:sec:swinl}
Swin Transformer, with its extensive design, offers advanced modeling capabilities, allowing for strong generalization across various tasks and datasets \cite{liu2021swin}.
In this paper, we employ its huge version (Swin-H) as the spatial feature extractor for HSROI in our SkySense.
To validate the impact of parameter size on model performance, we conduct experiments by replacing the Swin-H (654M) with Swin-L (197M), a variant with a smaller parameter size.
We perform experiments on three representative tasks (\ie, scene classification on the RESISC-45 dataset \cite{cheng2017remote}, object detection on the DIOR dataset \cite{li2020object}, and semantic segmentation on the Postdam dataset \cite{sherrah2016fully}), as shown in \cref{tab:smaller_param}.
Comparing with Swin-H, the adoption of Swin-L results in a reduction of the model's parameter count by 69.8\%.
The experimental results of downstream tasks show a slight decrease in performance when using Swin-L instead of Swin-H, thereby substantiating the effectiveness and necessity of employing the model with a larger parameter size, \ie, Swin-H.
We further compare our method equipping Swin-L with the two representative RSFMs: SatMAE \cite{cong2022satmae} and Scale-MAE \cite{reed2022scale}.
Importantly, the SkySense with Swin-L exhibits a significantly smaller parameter size (197M) compared to SatMAE (307M) and Scale-MAE (307M), while demonstrating a remarkably better performance in three downstream tasks.
In particular, on the RESISC-45 dataset with TR=10\%, SkySense with Swin-L outperforms SatMAE by 2.62\% and Scale-MAE by 1.71\% in OA.

Achieving the better performance with the smaller parameter size, our method demonstrates that its exceptional effectiveness is not simply scaling up model parameters. Instead, factors such as innovative network architecture design and advanced pre-training strategies also play vital roles. These factors, beyond the parameter size, significantly contribute to the exceptional performance of SkySense.

\begin{table*}[htbp]
\centering
\begin{tabular}{lccccc}
\toprule
\multirow{2}{*}{Model} & \multirow{2}{*}{Publication} & \multirow{2}{*}{\# Parameters} & RESISC-45                   & DIOR    & Postdam \\ \cmidrule(lr){4-6} 
                       &              &                   & OA (TR=10\% / 20\%) & $\mathrm{mAP}_{50}$ & mIoU  \\ \midrule
SatMAE \cite{cong2022satmae}     & NIPS'22            & 307M                            & 91.72 / 94.10         & 70.89   & 90.63 \\
Scale-MAE \cite{reed2022scale}    & ICCV'23           & 307M                            & 92.63 / 95.04         & 73.81   & 91.54 \\ \midrule
SkySense (Swin-L)       & -     & 197M                     & 94.34 / 95.92         & 76.74   & 92.86 \\
SkySense (Swin-H)   & -     & 654M                           & 94.85 / 96.32        & 78.73   & 93.99 \\ \bottomrule
\end{tabular}
\caption{Results of SkySense with smaller parameter size.}
\label{tab:smaller_param}
\end{table*}

\section{Comparison with random initialization and Vision Foundation Model}
\label{sup:sec:dino}
In this section, we employ both SkySense pre-trained weights and randomly initialized weights to fine-tune the same networks on 5 datasets over 4 different tasks. These tasks encompass scene classification using the AID dataset \cite{xia2017aid}, object detection employing the DIOR dataset \cite{li2020object}, semantic segmentation utilizing the Dyna.-S2 dataset \cite{toker2022dynamicearthnet} and iSAID dataset \cite{waqas2019isaid}, and change detection with the Dyna.-S2 dataset. The experimental results are presented in \cref{tab:dino}. The results across all 5 datasets demonstrate a substantial performance advantage of the our pre-trained model over the model learned from scratch. 

We further conduct experiments between SkySense and the DINOv2 \cite{oquab2023dinov2}, a well-established Vision Foundation Model, on Earth Observation interpretation tasks. Notably, DINOv2 is pre-trained on a large volume of natural images. The obtained experimental results manifest the conspicuous superiority of our method in all 5 datasets, surpassing the performance of DINOv2. We posit that the unsatisfactory performance of DINOv2 in remote sensing scenarios arises from two primary factors. Firstly, a discernible domain gap exists between remote sensing imagery and natural images, rendering it arduous for the DINOv2 model, pre-trained solely on natural images, to effectively generalize across the diverse data sources and modalities inherent in remote sensing interpretation tasks. Secondly, DINOv2 lacks specialized design considerations geared towards remote sensing's characteristics, particularly in terms of spatiotemporal knowledge of remote sensing imagery. Consequently, the model does not sufficiently leverage the abundant spatiotemporal knowledge from remote sensing imagery in the downstream tasks.
Conversely, SkySense is purposefully formulated for remote sensing tasks, with tailored pre-training data, methods, and model structures that align harmoniously with downstream remote sensing interpretation tasks. As a result, it exhibits a markedly superior performance.

\begin{table*}[htbp]
\centering
\begin{tabular}{lccccc}
\toprule
\multirow{3}{*}{Model} & Scene Classification    & Object Detection & \multicolumn{2}{c}{Semantic Segmentation} & Change Detection \\ \cmidrule(lr){2-2} \cmidrule(lr){3-3} \cmidrule(lr){4-5}\cmidrule(lr){6-6}  
                       & AID               & DIOR             & Dyna.-S2              & iSAID             & Dyna.-S2         \\ \cmidrule(lr){2-6} 
                       & OA (TR=20\%/ 50\%) & $\mathrm{mAP}_{50}$         & mIoU                  & mIoU              & SCS              \\ \midrule
Random Init. (from scratch)              & 65.56/90.57       & 55.16            & 25.7/36.0             & 47.89             & 14.0/15.9        \\
DINOv2 \cite{oquab2023dinov2}                 & 96.16/97.69       & 68.91            & 30.9/40.9             & 58.69             & 13.8/16.6        \\ \midrule
\textbf{SkySense }          & \textbf{97.68/98.60}       & \textbf{78.73}            & \textbf{33.1/46.2}             & \textbf{70.91}             & \textbf{15.4/18.0}        \\ \bottomrule
\end{tabular}
\caption{Results of the model learned from scratch, DINOv2 (ViT-L/14), and SkySense.}
\label{tab:dino}
\end{table*}

\section{Effectiveness of Factorized Spatiotemporal Encoder}
\label{sup:sec:factorized}
We conduct a validation of SkySense's fundamental unit, the Factorized Spatiotemporal (F-ST) encoder, and the results are presented in \cref{tab:model_archi}.
In this validation, we compare F-ST encoder with two alternative options: the 3D model (UNet3D \cite{m2019semantic}) and the factorized temporospatial model (TSViT \cite{tarasiou2023vits}). All three models are trained from scratch using Sentinel-2 data from the PASTIS-R dataset \cite{garnot2022multi} , with the purpose of evaluating the structure design only. To ensure fairness, we use the same training settings and a similar number of model parameters as described in \cite{tarasiou2023vits}. Results show our F-ST encoder exhibits superior performance with substantially fewer parameters than 3D structure. Moreover, compared with TSViT, our design enables much better flexibility, as the fusion component is flexible enough to accommodate dimensions beyond time, such as modality and knowledge.

\begin{table}
\centering
\begin{tabular}{lcc}
\toprule
\multirow{2}{*}{Architecture} & \multirow{2}{*}{\# Parameters} & PASTIS-R \\ \cmidrule(lr){3-3} 
                              &                             & OA       \\ \midrule
UNet3D \cite{m2019semantic}                       & 6.2M                        & 82.3     \\
TSViT \cite{tarasiou2023vits}                         & 1.7M                       & 83.4     \\ \midrule
\textbf{*F-ST encoder}                  & \textbf{2.3M}                        & \textbf{83.7}     \\ \bottomrule
\end{tabular}
\caption{Results of different space-time data processing architectures. Three models are trained from scratch using Sentinel-2 data from the PASTIS-R dataset. * denotes the adoption of an identical F-ST architecture as employed by our SkySense.}
\label{tab:model_archi}
\end{table}

\section{Effectiveness of Cross-Modal Alignment}
\label{sup:sec:alignment}
The Cross-Modal Alignment of SkySense plays a role in explicitly aligning features from different modalities, facilitating cross-modal interactions.
To intuitively validate the effectiveness of Cross-Modal Alignment, we conduct a feature visualization experiment. 
Specifically, we calculate the attention map of the output tokens of each Transformer layer in the Multi-modal Temporal Fusion Transformer.
Two cases of visualization experiments are conducted: one with the Cross-Modal Alignment before the Multi-modal Temporal Fusion Transformer, and the other without it.
The visualization results of different intermediate layers are shown in \cref{fig:unalign_align}.
The results in \cref{fig:unalign_align} (a) suggest that the model without Cross-Modal Alignment focuses their attention on the extra token and the token corresponding to HSROI. The unnecessary attention to the token corresponding to HSROI indicates that the information from the HSROI is not adequately integrated into the extra token, resulting in ineffective semantic information fusion.
After incorporating the Cross-Modal Alignment, the model predominantly concentrates on particular extra token as shown in in \cref{fig:unalign_align} (b), indicating successful integration of the tokens associated with HSROI, TMsI and TSARI within the model.
This indicates that the Cross-Modal Alignment indeed facilitates the comprehensive fusion of multi-modal RSI, which is crucial for constructing a Multi-Modal RSFM.

In addition to the results in Table 5(b) from our paper, we conduct an extra experiment on adding MAE into pre-training and a minor decrease (-0.7\% mIoU) is observed on the DynamicEarthNet test set. We argue that the pixel-level modeling of MAE is already addressed by our Multi-Granularity Contrastive Learning.

\section{Qualitative results of downstream tasks}
\label{sup:sec:visual}
We show qualitative results of SkySense and other representative RSFMs on various downstream tasks in this section. Specifically, we provide the visualizations of results and features on tasks including semantic segmentation, object detection, scene classification, \etc.
These visualizations are utilized to provide a more comprehensive and intuitive assessment of SkySense's performance.

\begin{figure*}[htbp]
    \centering
    \includegraphics[width=0.8\linewidth]{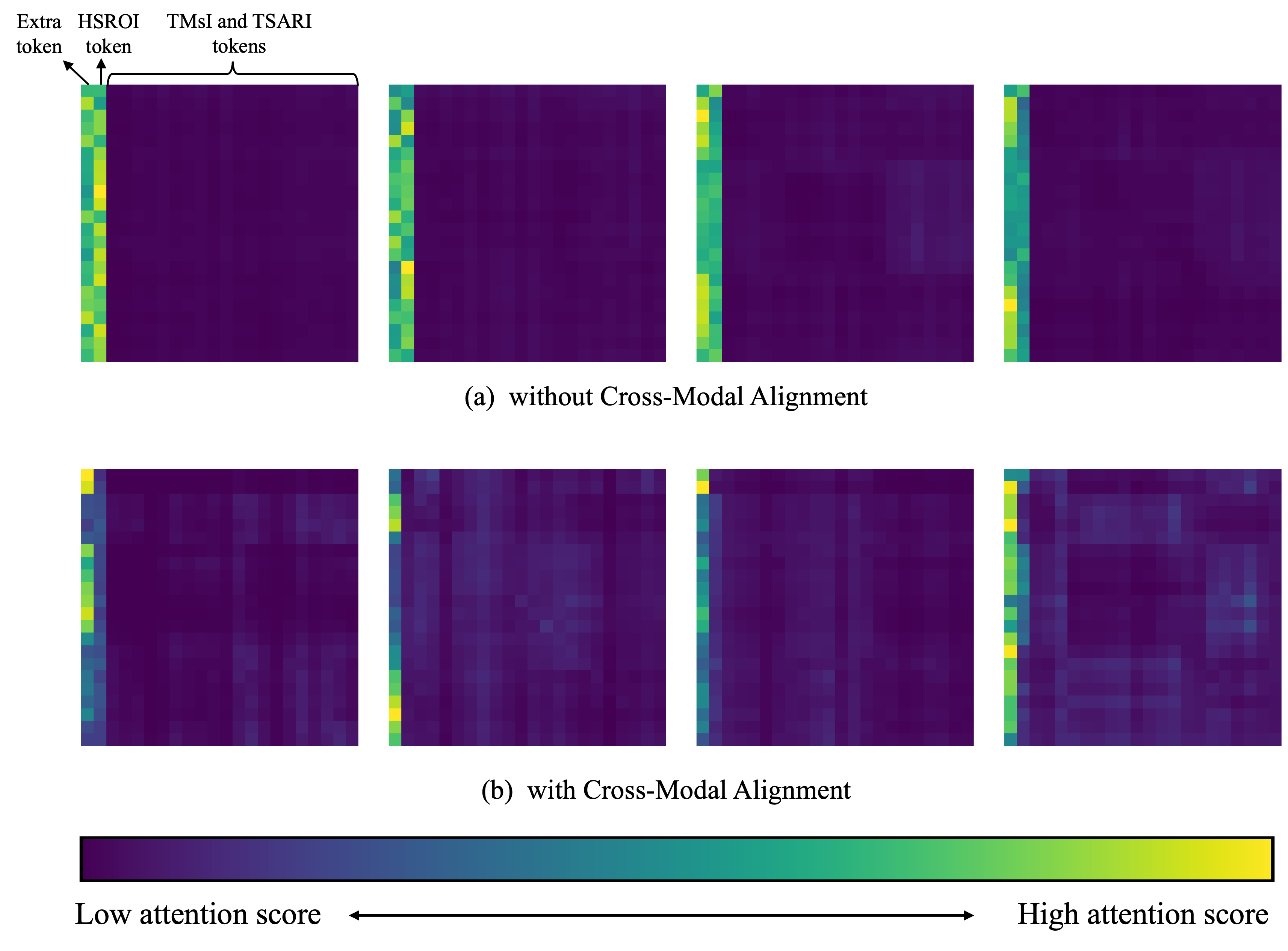}
    \caption{Attention map of the output tokens of Transformer layer in the Multi-modal Temporal Fusion Transformer. In each attention map, the first column corresponds to the extra token described in our paper, the second column represents the HSROI token, and the subsequent columns indicate TMsI and TSARI tokens. (a) Attention maps without the Cross-Modal Alignment; (b) correspondent attention maps with the Cross-Modal Alignment. The absence of Cross-Modal Alignment results in an unwarranted concentration of attention on the HSROI token, highlighting the inadequate integration for HSROI.}
    \label{fig:unalign_align}
\end{figure*}

\begin{figure*}[htbp]
  \centering
  \includegraphics[width=1.0\textwidth]{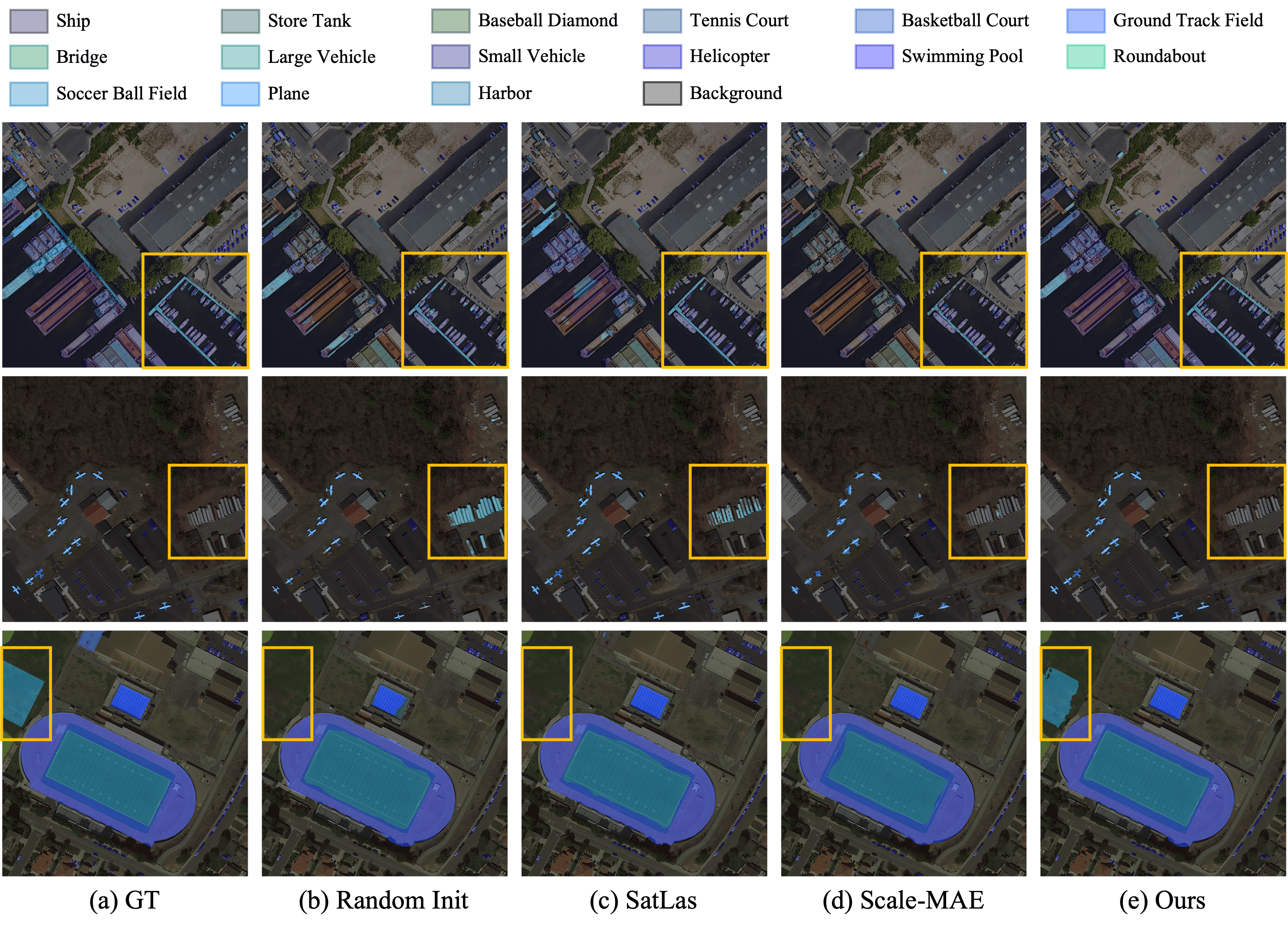}
  \caption{Visualization of semantic segmentation results on the iSAID dataset. }
  \label{fig:vis_isaid}
\end{figure*}

\noindent\textbf{Semantic Segmentation.}
We visualize the semantic segmentation results on the commonly used iSAID dataset \cite{waqas2019isaid}, as depicted in \cref{fig:vis_isaid}.
In the first row, the scene depicts a port area. In comparison to two recent methods, Scale-MAE \cite{reed2022scale} and SatLas \cite{bastani2022satlas}, SkySense achieves the most accurate segmentation results esspecially at \texttt{harbors} and \texttt{ships} across different scales.
We speculate that our model enhances its ability to segment multi-scale objects through Multi-Granularity Contrastive Learning.
The second row depicts an airport where planes are relatively small in terms of GSD. Other methods, particularly Scale-MAE, exhibit rough outlines in the regions of \texttt{plane} wings. Furthermore, both SatLas and Scale-MAE erroneously identify the containers in the right of the image as \texttt{large vehicles}.
Our SkySense consistently achieves the most outstanding results, demonstrating excellence both in overall segmentation and fine details.
Specifically, our method performs exceptionally well in segmenting \texttt{plane}, particularly in capturing the boundaries.
The scene in the third row features a sports stadium that includes a \texttt{ground track field}, a \texttt{swimming pool}, and a \texttt{soccer ball field}.
The \texttt{soccer ball field} on the left is not complete, posing challenges for accurate segmentation. Among various methods, SkySense successfully detects a large area of the \texttt{soccer ball field} and maintains a high consistency with the ground truth. In contrast, methods like SatLas and Scale-MAE even completely disregard the \texttt{soccer ball field} and provide incomplete predictions.
Throughout all the obtained results, our method consistently delivers superb visual results, which further demonstrates the effectiveness of our approach in downstream semantic segmentation tasks.

\begin{figure}[htbp]
  \centering
  \includegraphics[width=0.48\textwidth]{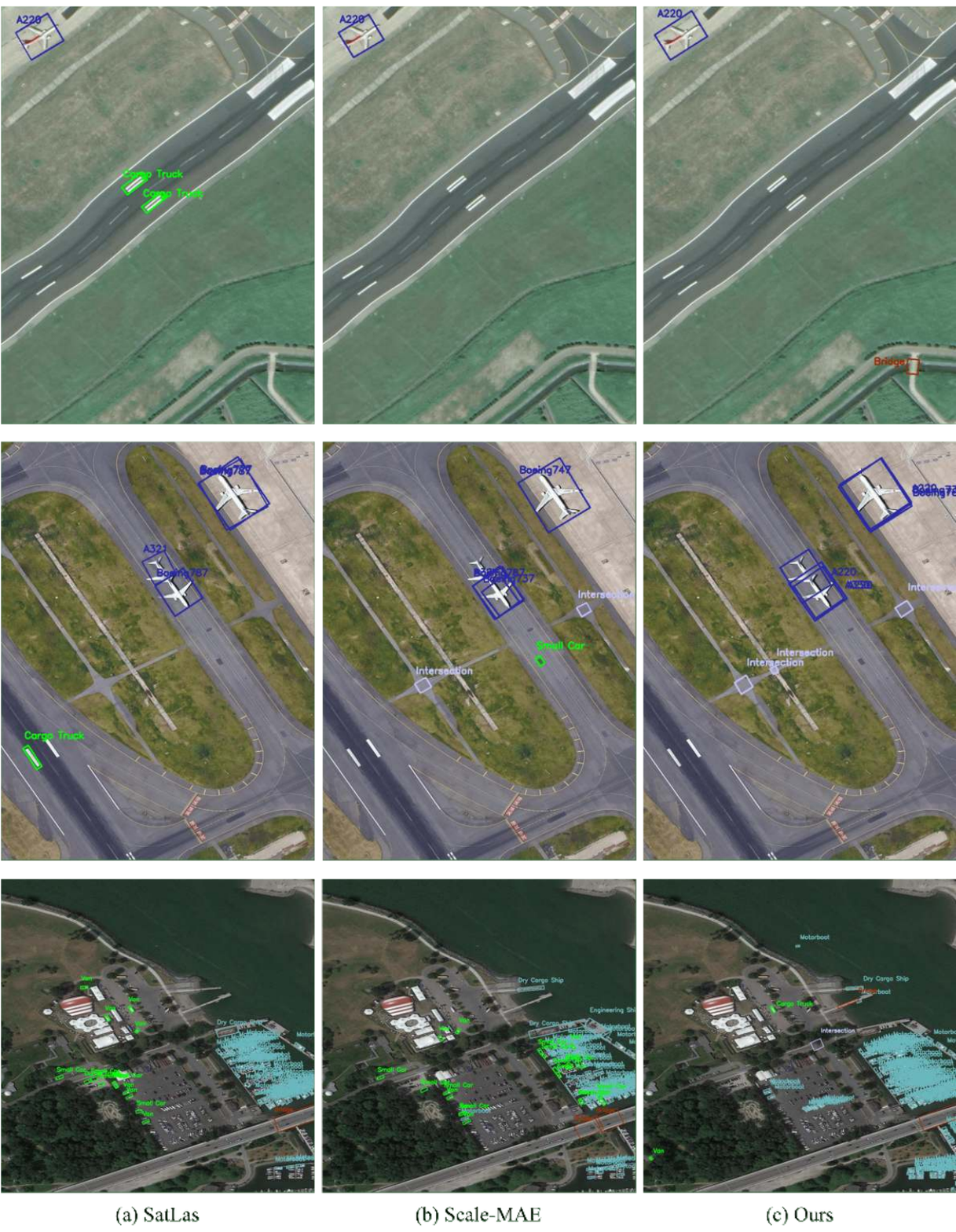}
  \caption{Visualization of object detection results on the FAIR1M dataset. It is worth noting that the labels for the test set have not been made public, therefore the visualization results do not contain ground truth.}
  \label{fig:vis_fair1m}
\end{figure}
\noindent\textbf{Object Detection.}
The object detection results on the FAIR1M dataset \cite{sun2022fair1m} are visualized in \cref{fig:vis_fair1m}.
It is noteworthy that the FAIR1M dataset does not disclose labels associated with the test set. Quantitative results are obtained through online evaluation\footnote{\url{https://www.gaofen-challenge.com/benchmark}}.
The first two rows of the visualized results feature airport scenes. In the first row, SatLas mistakenly identifies the \texttt{airplane runway lines} as \texttt{trucks}. Although SatLas and Scale-MAE can detect the \texttt{plane}, their bounding boxes do not accurately match the plane.  Parts of the \texttt{airplanes} are even outside the bounding boxes. Our method not only detects \texttt{planes} but also provides accurate bounding boxes for them.
The scene in the second row is noticeably more complex, with multiple \texttt{intersections} and \texttt{airplanes} that visually overlap.
Although SatLas captures all the \texttt{airplanes}, similar to the result in the first row, its bounding boxes do not accurately match the \texttt{airplanes}. Furthermore, in this case, SatLas mistakenly identifies an \texttt{airplane runway line} as a \texttt{truck} and completely overlooks \texttt{intersections}.
Scale-MAE performs well in detecting \texttt{airplanes}, but it misses one \texttt{intersection} in this scene and incorrectly detects the shadows on the \texttt{airplane runway} as \texttt{small cars}.
The object detection results presented above clearly indicate that SkySense surpasses other recent methods and achieves superior overall performance. It demonstrates exceptional generalization capabilities for downstream object detection task.

\begin{figure*}[htbp]
  \centering
  \includegraphics[width=0.96\textwidth]{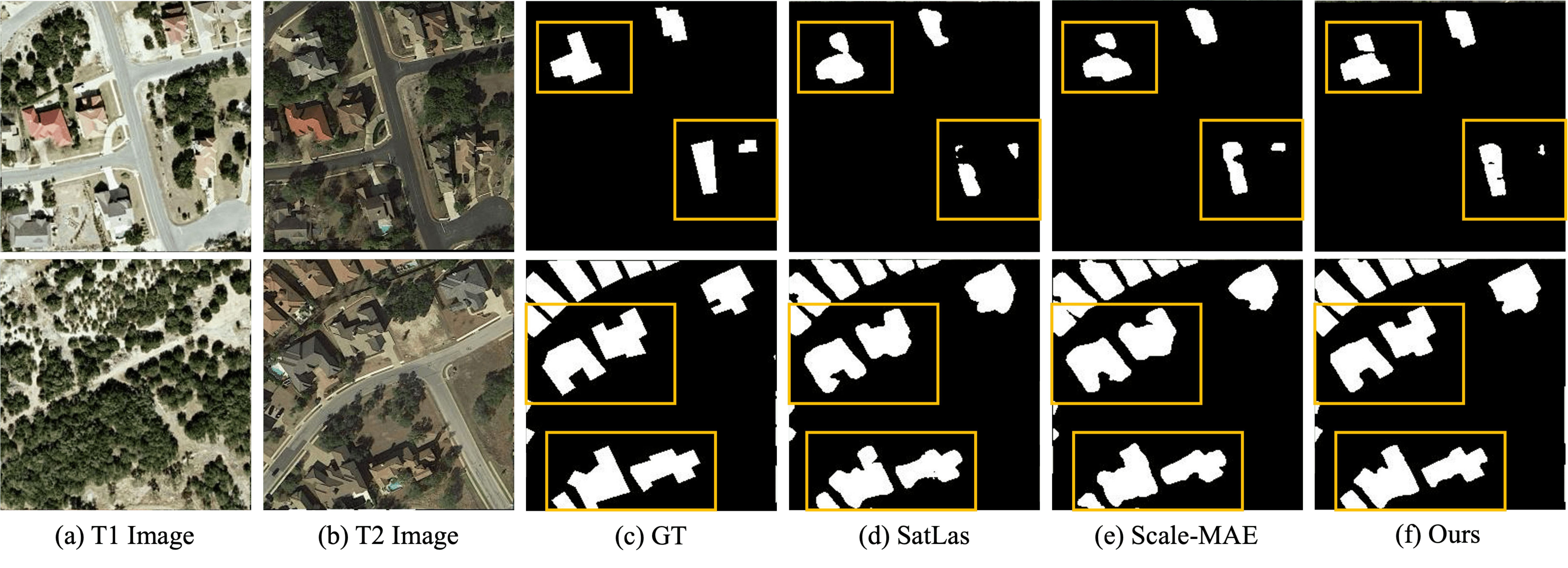}
  \caption{Visualization of change detection results on the LEVIR dataset. }
  \label{fig:vis_levir}
\end{figure*}
\noindent\textbf{Change Detection.}
We present the change detection visualization results obtained from the LEVIR-CD dataset \cite{chen2020spatial} as shown in \cref{fig:vis_levir}.
In the first row, the main changes between the bi-temporal images are the newly built buildings surrounding the existing ones. SatLas and Scale-MAE exhibit overly rough change boundaries. SkySense demonstrates the best visual results that matches the ground truth well.
The second row shows a significant increase in the number of buildings in the second temporal image. All three methods detect all the changed areas, and SkySense provides the best results regarding the boundaries.
The results mentioned above demonstrate that our SkySense, when used for change detection task, not only accurately identifies the changed objects but also provides precise and detailed boundaries.

\begin{figure*}[htbp]
  \centering
  \includegraphics[width=0.96\textwidth]{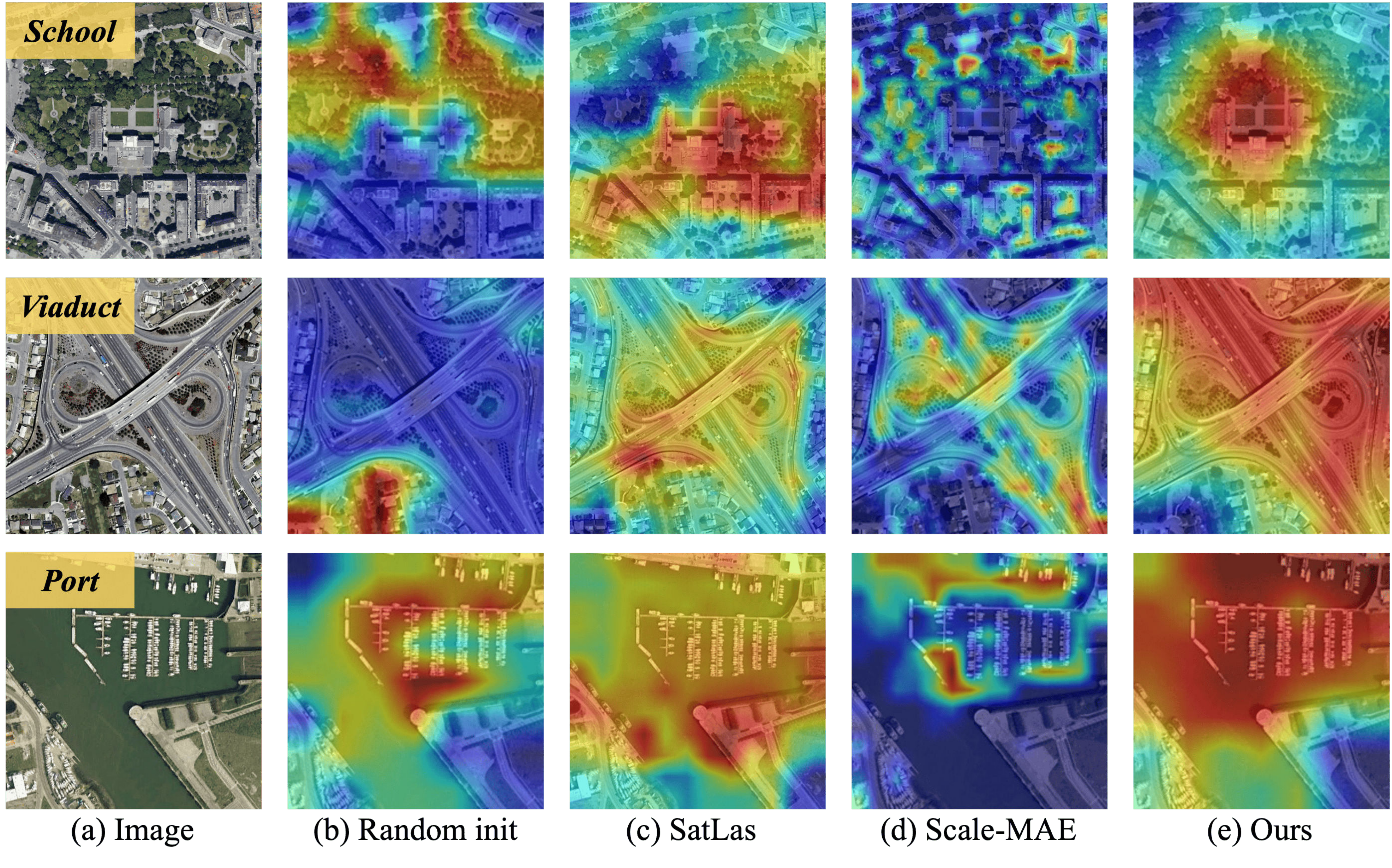}
  \caption{Visualization of Grad CAM on the AID dataset (TR=20\%). }
  \label{fig:vis_aid}
\end{figure*}
\noindent\textbf{Scene Classification.}
In \cref{fig:vis_aid}, we visualize the Gradient-weighted Class Activation Mapping (Grad CAM) \cite{selvaraju2017grad} on the AID dataset \cite{xia2017aid}, illustrating the superior feature learning capabilities of SkySense.
For the \texttt{school} category in the first row, SatLas exhibits an excessive emphasis on vegetation-covered areas, while Scale-MAE mainly focuses on some surrounding tiny buildings. SkySense effectively focuses on the main teaching building in the center of the scene and other associated areas, demonstrating its superior feature for this example.
The second row represents a \texttt{viaduct} scene. SatLas inadequately assigns attention scores to some of the \texttt{viaduct} region. Scale-MAE only focuses on certain local areas, significantly lacking focus on the main subjects. Our method effectively and accurately prioritizes all viaducts in the image, assigning them high attention scores.
In the third row, representing a \texttt{port} area, the observations bear a resemblance to those of the second row. Scale-MAE exhibits an excessive inclination towards smaller objects, while SatLas fails to assign an adequately high attention score to the primary subjects of interest.
Our method shows the most compelling visualization results.
From the above feature visualizations, it is evident that SkySense demonstrates better performance compared to Scale-MAE and SatLas. Scale-MAE tends to overly emphasize tiny objects due to its scale-aware design, while SatLas exhibits inadequate attention towards the primary targets, diminishing their performance in this classification task. In contrast, our SkySense's activation map effectively captures targets and their relevant features.

\begin{figure*}[htbp]
  \centering
  \includegraphics[width=0.96\textwidth]{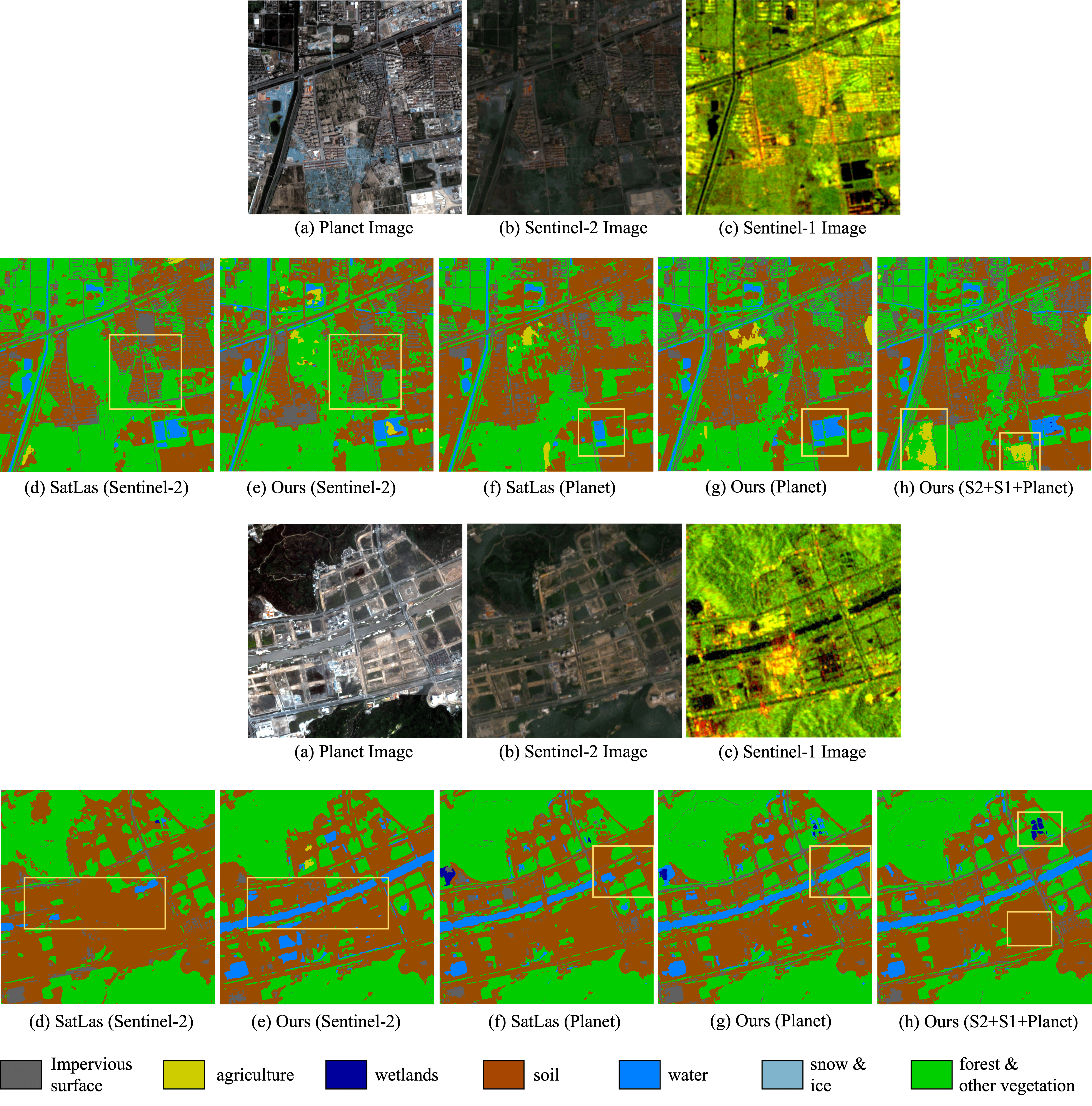}
  \caption{Visualization of multi-modal semantic segmentation results on the DynamicEarthNet-MM dataset. It is worth noting that the labels for the validation and test sets have not been made public, therefore the visualization results do not contain ground truth.}
  \label{fig:vis_dynamic}
\end{figure*}

\noindent\textbf{Multi-Modal Semantic Segmentation.}
One of the key advantages of SkySense is its support for multi-modal remote sensing imagery.
To intuitively validate the robustness and effectiveness of SkySense in multi-modal Earth Observation interpretation tasks, we conduct a visual analysis of the results of multi-modal semantic segmentation on the DynamicEarthNet-MM dataset \cite{toker2022dynamicearthnet}. 
This dataset provides diverse and comprehensive multi-modal remote sensing imagery, including PlanetFusion, Sentinel-1, and Sentinel-2 imagery, allowing us to examine the performance of our method across different modalities.
It is worth noting that the labels for the validation and test sets have not been made public, therefore the visualization results do not contain ground truth.
The qualitative results are illustrated in \cref{fig:vis_dynamic}.
The input images corresponding to the first case are shown in the first row of \cref{fig:vis_dynamic}. This particular scene provides a broad field of view, encompassing multiple land cover categories such as \texttt{agriculture}, \texttt{soil}, and \texttt{water}.
To provide a comprehensive comparison with SatLas, which does not support multi-modal data input, we also contrast the results of single-modal segmentation on Sentinel-2 and PlanetFusion images. 
Regarding the single-modal input of Sentinel-2 imagery, SkySense exhibits a higher accuracy in detecting small water bodies within the soil, primarily attributed to the incorporation of Multi-Granularity Contrastive Learning.
In the case of single-modal input from PlanetFusion imagery, SkySense successfully segments the rectangular water body area in the lower right corner, demonstrating better single-modal segmentation capabilities. SatLas fails to recognize those specific areas.
When the training data includes three sources: Sentinel-2, Sentinel-1, and PlanetFusion, SkySense exhibits segmentation performance that surpasses the single-modal segmentation. Specifically, with the advantage of multi-modal remote sensing imagery training, SkySense can successfully segment certain agricultural areas that are difficult to detect under single-modal conditions as shown in \cref{fig:vis_dynamic} (h).
The second case is shown in the third row of \cref{fig:vis_dynamic}. The input scene contains a river that runs through the entire image. SatLas, regardless of whether trained on Sentinel-2 or PlanetFusion data, fails to segment the continuous river.
Even when training on single-modal data, SkySense is capable of segmenting relatively continuous and complete rivers. However, there are some minor flaws in the details. For example, in the lower right portion of the image, \texttt{soil} is mistakenly detected as \texttt{water}, and the segmentation of \texttt{wetlands} in the upper right portion is incomplete.
When training on multiple modalities, SkySense shows significant visual improvement in this case. The segmentation flaws in the single-modal cases mentioned above are no longer present.
Thanks to the design of the Factorized Multi-modal Spatiotemporal Encoder that separates spatial feature extraction from multi-modal temporal fusion, Skysense can flexibly utilize either a single modality or multiple modalities for training, which gives it a notable advantage over existing methods. 
Additionally, in the aforementioned visualized results, even trained and tested with a single modality,  SkySense's performance still surpasses other methods, demonstrating the strong generalization capability and learning ability. With the training data of multiple modalities, the effectiveness of SkySense is further improved. This not only highlights the significant enhancement in semantic segmentation of remote sensing imagery by incorporating multiple modalities, but also underscores the importance of supporting multi-modal remote sensing imagery for RSFM.

After conducting a comprehensive analysis of the results, it becomes apparent that SkySense consistently outperforms other methods in a range of downstream tasks. Notably, it demonstrates remarkable accuracy in executing segmentation, detection, and classification processes. These findings effectively highlight the exceptional learning capacity and generalization capability of SkySense for downstream tasks.

\section{Pre-training dataset}
\label{sup:sec:dataset}
\begin{figure*}[htbp]
	\centering
		\includegraphics[scale=.045]{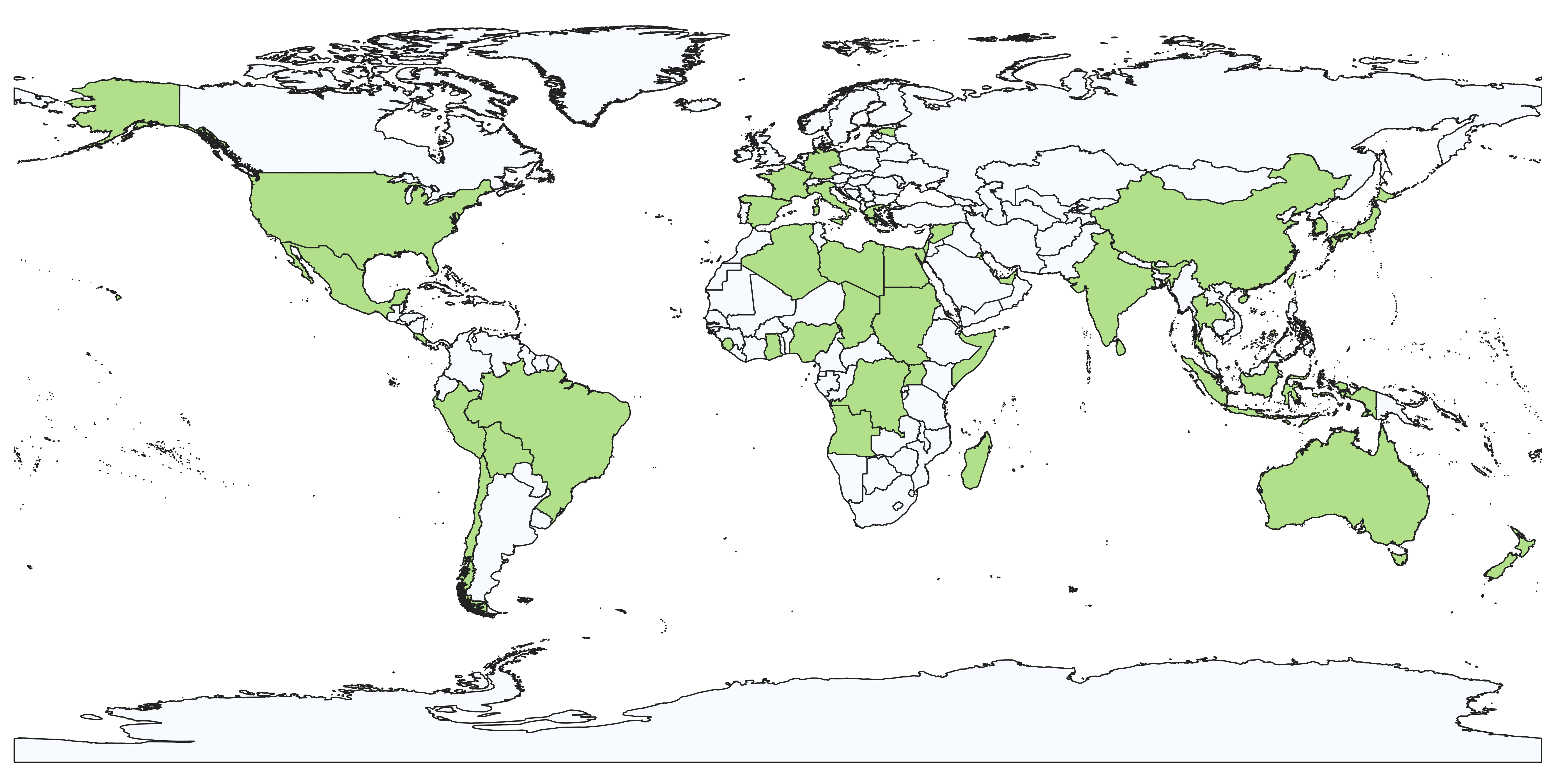}
	    \caption{Geographical distribution of pre-training data. The green areas represent the countries or regions covered by the pre-training data. The basic world map is obtained from \url{https://www.resdc.cn/data.aspx?DATAID=205.}}
        
	\label{fig:figr1}
\end{figure*}

\begin{figure*}[htbp]
	\centering
		\includegraphics[scale=.092]{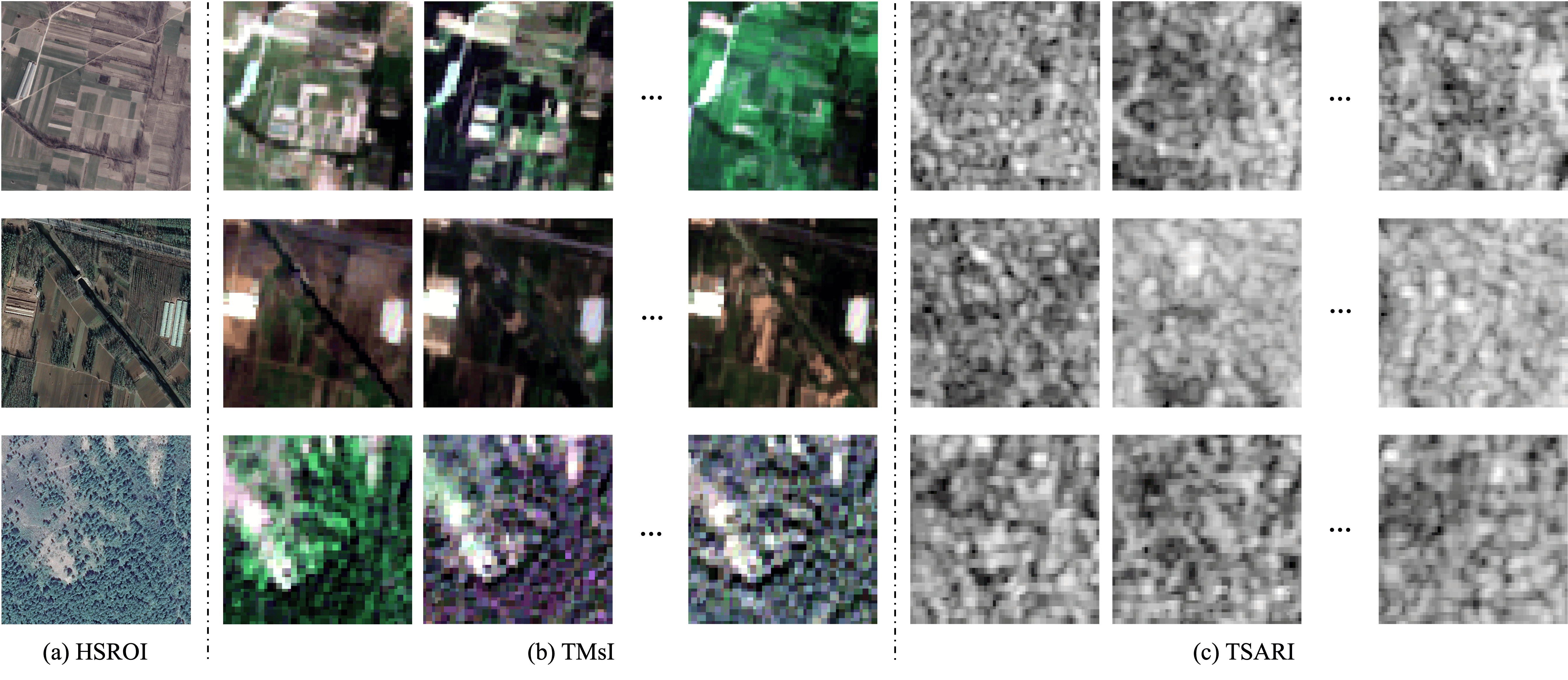}
	    \caption{Visualization of three training samples from our pre-training dataset. Each sample contains a HSROI, TMsIs and TSARIs. The first column represents HSROI. The second to fourth columns stand for multispectral false-color images captured at different times. The fifth to seventh columns are SAR grayscale images captured at different times.}

	\label{fig:figr2}
\end{figure*}

Existing remote sensing datasets lack the numerous amounts of multi-modal time-series Remote Sensing Imagery required for building SkySense, thus we develop a comprehensive multi-modal remote sensing dataset with temporal sequences specifically for SkySense pre-training.

\noindent{\textbf{Data Acquisition and Preprocessing.}}
This dataset comprises diverse sources of remote sensing imagery collected globally (see \cref{tabr1}), including HSROIs from WorldView-3, 4, etc., TMsI from Sentinel-2, and TSARI from Sentinel-1. 
\begin{itemize}
\item\textit{HSROIs.} We collect high-resolution optical RGB images from a third-party platform, with an average Ground Sample Distance (GSD) of 0.3 meter.

\item\textit{TMsI.} We collect the freely available Sentinel-2 level-2A atmospherically corrected surface reflectance sequence images as another significant data source. The bands with a resolution of 10m (visible and NIR) and resampled bands with a resolution of 20m (Vegetation Red Edge and SWIR) are merged to form a multispectral image (10 bands). In addition, cloudy imagery filtering is implemented to obtain higher quality multispectral data. Specifically, according to the Scene Classification Map\footnote{\url{https://custom-scripts.sentinel-hub.com/custom-scripts/sentinel-2/scene-classification/}} provided by the European Space Agency, images with a cloud coverage ratio exceeding 1\% (calculated as the sum of the proportion of pixels belonging to the categories of \texttt{Thin cirrus}, \texttt{Cloud medium probability}, \texttt{Cloud high probability}, and \texttt{Cloud shadows} categories) are filtered out.

\item\textit{TSARI.} We obtain the easily accessable Sentinel-1 ground-range-detected (GRD) products with both VV and VH polarization, which are acquired in the interferometric wide swath (IW) mode. And the standard calibration\footnote{\url{https://sentinels.copernicus.eu/web/sentinel/toolboxes/sentinel-1}} process is employed to obtain the  spatially-aligned SAR images. These images contain the backscatter coefficient ($\sigma^{\circ}$) in decibels (dB).

\end{itemize}

\noindent{\textbf{Dataset statistics.}}
As shown in \cref{fig:figr1}, our dataset spans over 8.78 million square kilometers across 40 countries and areas. It covers 6 continents, \ie, Asia, Europe, Africa, North America, South America and Oceania. The dataset contains 21.5 million training sequences, each consisting of 1 static HSROI, a randomly-sampled TMsI of sequence length 20 and a randomly-sampled TSARI of sequence length 10. In total it occupies a storage space of around 300 Terabytes. Our dataset exhibits significant complementarity in terms of temporal information, spatial resolution, and imaging mechanisms, given the three modalities.

\begin{table}[htb]
\centering
\setlength\tabcolsep{2pt}%
\scriptsize
\begin{tabular}{cccccc}
\toprule
\textbf{Modality}  & \textbf{Sensor}                                                                  & \textbf{Band / Polarization}                                                        & \textbf{\begin{tabular}[c]{@{}c@{}}GSD \\ (m)\end{tabular}} & \textbf{\begin{tabular}[c]{@{}c@{}}Image \\ Size (px)\end{tabular}} & \textbf{\begin{tabular}[c]{@{}c@{}}Avg. seq \\ Length\end{tabular}} \\ \midrule
\begin{tabular}[c]{@{}c@{}}Optical \\ (HSROI)\end{tabular} & \begin{tabular}[c]{@{}c@{}}WorldView-3,4, \\ etc.\end{tabular}           & RGB                                                         & 0.3                                                & 2048 $\times$ 2048                                           & 1                                                          \\ \midrule
\begin{tabular}[c]{@{}c@{}}Optical \\ (TMsI)\end{tabular}  & \begin{tabular}[c]{@{}c@{}}Sentinel-2 \\ (Level-2A)\end{tabular}        & \begin{tabular}[c]{@{}c@{}}B2-8, B8A,\\ B11-12\end{tabular} & 10                                                 & 64 $\times$ 64                                               & 65                                                         \\ \midrule
\begin{tabular}[c]{@{}c@{}}SAR \\ (TSARI)\end{tabular}     & \begin{tabular}[c]{@{}c@{}}Sentinel-1 \\ (Level-1 IW, GRD)\end{tabular} & VV, VH                                                      & 10                                                 & 64 $\times$ 64                                               & 13                                                         \\ \bottomrule
\end{tabular}
\caption{Statistics of our pre-training dataset. Ground sample distance (GSD) is the distance between the center of one pixel to the center of an adjacent pixel in a remote
sensing image.}
\label{tabr1}
\end{table}

\noindent{\textbf{Example visualization.}} \cref{fig:figr2} illustrates some examples from our pre-training dataset. Each example consists of a HSROI, TMsIs and TSARIs.

\section{Pre-training implementation details}
\label{sup:sec:implementation}
SkySense is pre-trained with a batch size of 240 samples for a total of 875k steps, distributed over 80 A100-80GB GPUs with the AdamW optimizer~\cite{loshchilov2018fixing}. We adopt a learning rate warmup~\cite{goyal2017accurate}, followed by a decay using a cosine schedule~\cite{loshchilov2016sgdr} from 0.04 to 0.2. For HSROIs, we apply augmentations including multi-crop~\cite{caron2020unsupervised}, Gaussian blur, solarization~\cite{grill2020bootstrap}, \etc. As for TMsI and TSARI, we randomly select two fixed-sized sequences (\ie 20 for TMsI and 10 for TSARI) from the original ones and perform random disturbances on the RSI acquisition date. We employ the huge version\footnote{\url{https://github.com/open-mmlab/mmpretrain}} of the Swin Transformer (Swin-H)~\cite{liu2021swin} as the spatial encoder for HSROIs, chosen for its design efficiency in minimizing computational costs for high-resolution imagery~\cite{zhang2021multi}. RSI from TMsI or TSARI is equipped with a ViT-L~\cite{dosovitskiy2020image}. The Multi-Modal Temporal Fusion Transformer contains 24 Naive Transformer encoder layers. For Geo-Context Prototype Learning, we divide the globe into 4096 regions. One region covers approximately 4294 square kilometers area and contains 100 prototypes. SkySense comprises a total of 2.06 billion parameters, specifically 654 million from the Swin-H and 302 million from a single ViT-L, as shown in \cref{parameter}.

The pixel size of our pre-training data is shown in \cref{tabr1}, which is up to 2048 $\times$ 2048. The pre-training takes 24600 A100 GPU hours. Its computational complexity is 4488.69 GFLOPs.

\begin{table}
\centering
\small
\begin{tabular}{l|cc}
\toprule
\textbf{Model Modules}                                                                   & \textbf{Architecture}                                                             & \textbf{\# Parameters} \\ \midrule
Spatial Encoder-HSROI                                                              & Swin-Huge                                                                & 654M          \\
Spatial Encoder-TMsI                                                               & ViT-Large                                                                & 302M          \\
Spatial Encoder-TSARI                                                              & ViT-Large                                                                & 302M          \\
\begin{tabular}[c]{@{}l@{}}Multi-modal Temporal \\ Fusion Transformer\end{tabular} & \begin{tabular}[c]{@{}c@{}} Transformer \\ Encoder\end{tabular} & 398M          \\
Geo-Context Prototype                                                              & -                                                                        & 215M          \\
Others                                                                             & -                                                                        & 189M          \\ \bottomrule
\end{tabular}
\caption{Parameter breakdown for each module of SkySense.}
\label{parameter}
\end{table}

\begin{table*}[htbp]
    \begin{subtable}[t]{1\linewidth}
        \centering
        \setlength\tabcolsep{2pt}%
        \small
        \begin{tabular}{l|cccc|ccc}
        \hline
        Task                    & \multicolumn{4}{c|}{\texttt{(i)} Semantic Segmentation}                                                                                                                                                                                                                                                                                                  & \multicolumn{3}{c}{\texttt{(ii)} Object Detection}       \\ \hline
        Dataset                 & Dyna.-Pla.                                                    & iSAID                                                                                             & Potsdam                                                                                           & Dyna.-S2                                                      & DIOR         & DIOR-R       & FAIR1M       \\ \hline
        Optimizer               & AdamW                                                            & AdamW                                                                                             & AdamW                                                                                             & AdamW                                                            & AdamW        & AdamW        & AdamW        \\
        Input Size              & 1024$\times$1024                                                        & 896$\times$896                                                                                           & 512$\times$512                                                                                           & 256$\times$256                                                          & 800$\times$800      & 800$\times$800      & 512$\times$512      \\
        Input channel           & RGBNIR                                                           & RGB                                                                                               & NIRRG                                                                                             & \begin{tabular}[c]{@{}c@{}}B02-08, B8A, \\ B11-12\end{tabular}   & RGB          & RGB          & RGB          \\
        Base learning rate      & 6e-5                                                             & 6e-5                                                                                              & 6e-5                                                                                              & 6e-5                                                             & 1e-4         & 1e-4         & 1e-4         \\
        Learning rate scheduler & poly                                                             & poly                                                                                              & poly                                                                                              & poly                                                             & multistep         & multistep    & multistep         \\
        Weight decay            & 0.01                                                             & 0.01                                                                                              & 0.01                                                                                              & 0.01                                                             & 0.05         & 0.05         & 0.05         \\
        Batch size              & 8                                                                & 16                                                                                                & 16                                                                                                & 8                                                                & 4            & 2            & 12           \\
        Max iteration/epoch     & 8k iters                                                         & 80k iters                                                                                         & 80k iters                                                                                         & 80k iters                                                        & 12 epoch     & 12 epoch     & 8 epoch      \\
        Warmup                  & linear                                                           & linear                                                                                            & linear                                                                                            & linear                                                           & linear       & linear       & linear       \\
        Warmup iteration/epoch  & 1.5k iters                                                       & 1.5k iters                                                                                        & 1.5k iters                                                                                        & 1.5k iters                                                       & 1k Iters     & 1k iters     & 500 iters    \\
        Warmup ratio            & 1e-6                                                             & 1e-6                                                                                              & 1e-6                                                                                              & 1e-6                                                             & 1e-3         & 1e-3         & 1e-3         \\
        Drop path rate          & 0.3                                                              & 0.3                                                                                               & 0.3                                                                                               & 0.3                                                              & 0.3          & 0.3          & 0.3          \\
        Augmentation            & \begin{tabular}[c]{@{}c@{}}RandomCrop,\\ RandomFlip\end{tabular} & \begin{tabular}[c]{@{}c@{}}RandomScaling \\ (0.5 to 2.0),\\ RandomCrop,\\ RandomFlip\end{tabular} & \begin{tabular}[c]{@{}c@{}}RandomScaling \\ (0.5 to 2.0),\\ RandomCrop,\\ RandomFlip\end{tabular} & \begin{tabular}[c]{@{}c@{}}RandomCrop,\\ RandomFlip\end{tabular} & RandomFlip   & RandomFlip   & \begin{tabular}[c]{@{}c@{}}RandomFlip,\\ RandomRotate \\ Multi-Scale\end{tabular}   \\
        Head/Detector          & UperNet                                                              & UperNet                                                                                               & UperNet                                                                                               & UperNet                                                              & Faster RCNN          & Oriented RCNN          & Oriented RCNN           \\
        Loss function          & CrossEntropy                                                             & CrossEntropy                                                                                               & CrossEntropy                                                                                               & CrossEntropy                                                              & \begin{tabular}[c]{@{}c@{}}CrossEntropy,\\ SmoothL1 \end{tabular}          & \begin{tabular}[c]{@{}c@{}}CrossEntropy,\\ SmoothL1 \end{tabular}           &     \begin{tabular}[c]{@{}c@{}}CrossEntropy,\\ SmoothL1 \end{tabular}      \\\hline
        \end{tabular}
    \end{subtable}

    \begin{subtable}[t]{1.0\linewidth}
        \vspace{1em}
        \centering
        \setlength\tabcolsep{2.5pt}%
        \small
        \begin{tabular}{l|cccc|ccc}
        \hline
        Task                    & \multicolumn{4}{c|}{\texttt{(iii)} Scene Classification}                                                                                                                                                                                                                                   & \multicolumn{3}{c}{\texttt{(iv)} Change Detection}                                                                                                                                                                                   \\ \hline
        Dataset                 & AID                                                                 & RESISC-45                                                           & BEN-S2                                                         & fMoW-S2                                                        & LEVIR-CD                                                                       & OSCD                                                               & Dyna.-S2                                                         \\ \hline
        Optimizer               & AdamW                                                               & AdamW                                                               & AdamW                                                          & AdamW                                                          & AdamW                                                                          & Adam                                                               & AdamW                                                            \\
        Input Size              & 224$\times$224                                                             & 224$\times$224                                                             & 128$\times$128                                                        & 96$\times$96                                                          & 256$\times$256                                                                        & 96$\times$96                                                              & 256$\times$256                                                          \\
        Input channel           & RGB                                                                 & RGB                                                                 & \begin{tabular}[c]{@{}c@{}}B02-08, B8A, \\ B11-12\end{tabular} & \begin{tabular}[c]{@{}c@{}}B02-08, B8A, \\ B11-12\end{tabular} & RGB                                                                            & \begin{tabular}[c]{@{}c@{}}B02-08, B8A, \\ B11-12\end{tabular}     & \begin{tabular}[c]{@{}c@{}}B02-08, B8A, \\ B11-12\end{tabular}   \\
        Base learning rate      & 6.25e-5                                                             & 6.25e-5                                                             & 5e-5                                                           & 8e-4                                                           & 6e-5                                                                           & 6e-4                                                               & 6e-5                                                             \\
        Learning rate scheduler & \begin{tabular}[c]{@{}c@{}}Cosine\\ Annealing\end{tabular}          & \begin{tabular}[c]{@{}c@{}}Cosine\\ Annealing\end{tabular}          & MultiStepLR                                                    &      \begin{tabular}[c]{@{}c@{}}Cosine\\ Annealing\end{tabular}                                                          & LambdaLR                                                                       & ExponentialLR                                                      & poly                                                             \\
        Weight decay            & 0.05                                                                & 0.05                                                                & 0.01                                                           & 0.05                                                           & 0.01                                                                           & 1e-4                                                               & 0.05                                                             \\
        Batch size              & 64                                                                  & 64                                                                  & 256                                                            & 1024                                                           & 8                                                                              & 32                                                                 & 8                                                                \\
        Max iteration/epoch     & 200 epoch                                                           & 200 epoch                                                           & 100 epoch                                                      & 30 epoch                                                       & 200 epoch                                                                      & 100 epoch                                                          & 80k iters                                                        \\
        Warmup                  & linear                                                              & linear                                                              & -                                                              & linear                                                              & -                                                                              & -                                                                  & linear                                                           \\
        Warmup iteration/epoch  & 5 epoch                                                             & 5 epoch                                                             & -                                                              & 5 epoch                                                              & -                                                                              & -                                                                  & 1.5k iters                                                       \\
        Warmup ratio            & 0.01                                                                & 0.01                                                                & -                                                              & 0.2                                                              & -                                                                              & -                                                                  & 1e-6                                                             \\
        Drop path rate          & 0.2                                                                 & 0.2                                                                 & -                                                              & 0.2                                                            & -                                                                              & -                                                                  & 0.3                                                              \\
        Augmentation            & \begin{tabular}[c]{@{}c@{}}RandomCrop,\\ RandomErasing\end{tabular} & \begin{tabular}[c]{@{}c@{}}RandomCrop,\\ RandomErasing\end{tabular} & RandomFlip                                                              & \begin{tabular}[c]{@{}c@{}}RandomCrop, \\, RandomFlip, \\ Mixup,\\ CutMix\end{tabular}        & \begin{tabular}[c]{@{}c@{}}RandomCrop,\\ RandomFlip,\\ RandomBlur\end{tabular} & \begin{tabular}[c]{@{}c@{}}RandomFlip,\\ RandomRotate\end{tabular} & \begin{tabular}[c]{@{}c@{}}RandomCrop,\\ RandomFlip\end{tabular} \\
        Head/Detector          & \begin{tabular}[c]{@{}c@{}}Linear\\ Classifier\end{tabular}                                                              & \begin{tabular}[c]{@{}c@{}}Linear\\ Classifier\end{tabular}                                                                                               & \begin{tabular}[c]{@{}c@{}}Linear\\ Classifier\end{tabular}                                                                                               & \begin{tabular}[c]{@{}c@{}}Linear\\ Classifier\end{tabular}                                                              & BIT           &  U-Net         & UperNet           \\
        Loss function          & CrossEntropy                                                             & CrossEntropy                                                                                               & \begin{tabular}[c]{@{}c@{}}MultiLabel\\ SoftMargin\end{tabular}                                                                                               & \begin{tabular}[c]{@{}c@{}}SoftTarget\\ CrossEntropy\end{tabular}                                                                  & CrossEntropy          & BCE           & CrossEntropy \\\hline
        \end{tabular}
    \end{subtable}
    \caption{The finetuning setting in single-modal downstream tasks.}
    \label{singleimple}
\end{table*}

\begin{table*}[htbp]
\centering
\setlength\tabcolsep{2.5pt}%
\small
\begin{tabular}{l|c|c|c}
\hline
Task                    & \begin{tabular}[c]{@{}c@{}}\texttt{(i)} Multi-Modal Segmentation: \\ Time-insensitive LandCover Mapping\end{tabular} & \begin{tabular}[c]{@{}c@{}}\texttt{(ii)} Multi-Modal Segmentation: \\ Time-sensitive Crop Mapping\end{tabular} & \multicolumn{1}{c}{\begin{tabular}[c]{@{}c@{}} \texttt{(iii)} Multi-Modal \\ Classification\end{tabular}} \\ \hline
Dataset                 & Dyna.-MM                                                                                      & PASTIS-MM                                                                                        & BEN-MM                                                                         \\ \hline
Optimizer               & AdamW                                                                                                         & AdamW                                                                                                 &   AdamW                                                                                         \\
Input Size              &   \makecell[c]{planet: 1024$\times$1024  \\ sentinel2: 1024$\times$1024 \\ sentinel1: 1024$\times$1024}                                                                                                       &   \makecell[c]{gep: 4096$\times$4096 \\ sentinel2: 128$\times$128 \\ sentinel1: 128$\times$128}                                                                                               &     \makecell[c]{sentinel2: 128$\times$128 \\ sentinel1: 128$\times$128}                                                                                        \\
Input channel           &        \makecell[c]{planet: RGBNIR \\ sentinel2: B02-08, B8A, B11-12 \\ sentinel1: VV, VH}                                                                                                 &    \makecell[c]{gep: RGB \\ sentinel2: B02-08, B8A, B11-12 \\ sentinel1: VV, VH}                                                                                              &     \makecell[c]{sentinel2: B02-08, B8A, B11-12 \\ sentinel1: VV, VH}                                                                                       \\
Base learning rate      &      6e-05                                                                                                   &     6e-05                                                                                             &     5e-05                                                                                      \\
Learning rate scheduler &     linear                                                                                                    &      linear                                                                                            &     MultiStepLR                                                                                       \\
Weight decay            &       0.01                                                                                                  &     0.01                                                                                             &     0.01                                                                                        \\
Batch size              &         8                                                                                                &      8                                                                                            &     256                                                                                       \\
Max iteration/epoch     &        6k iters                                                                                                 &      20k iters                                                                                            &     100 epoch                                                                                      \\
Warmup                  &      linear                                                                                                   &     linear                                                                                             &    -                                                                                       \\
Warmup iteration/epoch  &        150 iters                                                                                                 &    1500 iters                                                                                              &       -                                                                                   \\
Warmup ratio            &      1e-6                                                                                                   &    1e-6                                                                                              &      -                                                                                   \\
Drop path rate          &          0.3                                                                                               &      0.3                                                                                            &    -                                                                                       \\
Augmentation            &       \makecell[c]{RandomFlip}                                                                                                  &      \makecell[c]{RandomFlip}                                                                                            &            \makecell[c]{RandomFlip}                                                                                \\
Head/Detector            &       UperNet                                                                                                 &      FCN                                                                                            &     Linear Classifier                                                                                       \\
Loss function            &  CrossEntropy                                                                                                      &  CrossEntropy                                                                                                &   \makecell[c]{MultiLabel \\ SoftMargin}                                                                                  \\\hline
\end{tabular}
\caption{The finetuning setting in multi-modal downstream tasks.}
\label{multimodalimple}
\end{table*}

\section{Dataset and implementation details of downstream tasks}
\label{sup:sec:dowmstream}
In this section, we introduce the experimental datasets and implementation details used in downstream task.

\noindent\textbf{Semantic Segmentation.} Semantic segmentation serves as a prevalent application in remote sensing, facilitating the automatic extraction of land use classes and ground instances. Considering factors such as spatial resolution, spectrum and number of categories, we select four popular datasets for the semantic segmentation task:
\begin{enumerate}[label=\arabic*)]
\item\textit{DynamicEarthNet-PlanetFusion (Dyna.-Pla.) \cite{toker2022dynamicearthnet}.} The dataset comprises a collection of images from 75 global locations captured from the PlanetFusion satellite platform. The image acquisition period spans from January 2018 to December 2019. Each location has 24 images and corresponding annotations of 7 land use and land cover semantic classes. Each image contain four bands, namely Red, Green, Blue and Near-Infrared, with a GSD of 3 meters and an image size of 1024$\times$1024. Based on the official leaderboard\footnote{\url{https://codalab.lisn.upsaclay.fr/competitions/2882\#results}}, these locations are divided into 55 for training, 10 for validation, and 10 for testing. In the experiment, we use the official validation and test sets for evaluation. It is worth noting that ground truth labels for both the validation and test sets are unavailable for local users. Therefore, we submit the predictions to online leaderboard and report the obtained scores.

\item\textit{iSAID \cite{waqas2019isaid}.} This dataset comprises 2806 remote sensing images with dense annotations from multiple satellite sensors. The images vary in size from 800$\times$800 to 4000$\times$13000 pixels. It includes pixel-level annotations of 655451 instances across 15 object categories, while the remaining non-object pixels are labeled as \texttt{background}. It has been divided into training, validation, and test sets, consisting of 1411, 458, and 937 samples, respectively. Following \cite{sun2022ringmo,wang2022advancing}, the performance evaluation of RSFMs is conducted on the validation set.

\item\textit{Potsdam \cite{sherrah2016fully}.} A total of 38 aerial images with a GSD of 0.05 meters are collected to form the Potsdam dataset. These images are divided into 24 training images and 14 testing images. Each image has a fixed pixel size of 6000$\times$6000. Following \cite{sun2022ringmo}, we utilize images composed of Near-Infrared, Red, and Green spectral bands. The evaluation is conducted on the test set, focusing on five categories: impervious surfaces, buildings, low vegetation, trees, and cars. It is important to note that the clutter category is not included in the evaluation.

\item\textit{DynamicEarthNet-Sentinel2 (Dyna.-S2) \cite{toker2022dynamicearthnet}.} This dataset can be viewed as the Sentinel-2 data version of the abovementioned DynamicEarthNet-PlanetFusion dataset. Specifically, the DynamicEarthNet-Sentinel2 dataset offers monthly Sentinel-2 multispectral images, acquired between January 2018 and December 2019, that are spatially aligned with the corresponding PlanetFusion images. Each image in the dataset comprises 12 spectral channels and is uniformly resampled to match the sizes of PlanetFusion images, which are 1024$\times$1024 pixels. Following the same official data split protocol, we report the mIoU metric evaluated on the leaderboard-val and -test in our paper.

\end{enumerate}

We employ the UperNet \cite{xiao2018unified} as the unified segmentation head based on MMSegmentation\footnote{\url{https://github.com/open-mmlab/mmsegmentation}}, following \cite{sun2022ringmo, cha2023billion, wang2022advancing}. The detailed fine-tuning setting can be found in \cref{singleimple} (\texttt{i}).

\noindent\textbf{Horizontal \& Oriented Objection Detection.}
We employ the DIOR dataset to evaluate the performance of SkySense and other RSFMs for horizontal object detection task. Following \cite{sun2022ringmo}, we utilize Faster RCNN \cite{ren2015faster} as the detector, and other details are presented on \cref{singleimple} (\texttt{ii}).
\begin{enumerate}[label=\arabic*)]
\item \textit{DIOR \cite{li2020object}.} This dataset includes 23463 visible remote sensing images and 192472 object instances, which are manually annotated with horizontal bounding boxes and categorized into 20 common object classes. The image size in the dataset is of 800$\times$800 pixels, with GSD ranging from 0.5 meters to 30 meters. The dataset is divided into a training set consisting of 5862 patches, a validation set comprising 5863 patches, and a test set totaling 11738 patches. Following \cite{sun2022ringmo}, we mix the training set and validation set during training, while the test set is reserved for evaluation. In particular, its high inter-class similarity and intra-class diversity make precise localization and classification extremely challenging.
\end{enumerate}

Remote sensing images encompass a wide range of objects, such as buildings, vehicles, bridges and so on. These objects are densely distributed and display variations in terms of size, scale and orientation. Consequently, detecting and identifying these objects presents a significant challenge, commonly referred to as oriented object detection \cite{wen2023comprehensive}. To assess the performance of RSFMs on this task, we utilize the DIOR-R and FAIR1M datasets and employ the Oriented RCNN \cite{li2022oriented} as the detector, following \cite{sun2022ringmo,wang2022advancing,cha2023billion}. The specific implementation details can be found in \cref{singleimple} (\texttt{ii}) as well.

\begin{enumerate}[label=\arabic*)]
\setcounter{enumi}{1}
\item \textit{DIOR-R \cite{cheng2022anchor}.} The DIOR-R dataset shares the same images as the abovementioned DIOR dataset. However, the difference lies in the annotated oriented bounding boxes, which make it suitable for oriented object detection task. Similar to the implementation on the DIOR dataset, we follow \cite{wang2022advancing} by merging the training and validation sets during the training process, while the test set is reserved for evaluation.

\item \textit{FAIR1M \cite{sun2022fair1m}.} FAIR1M is a large-scale benchmark dataset for fine-grained oriented object detection, including over 40000 high-resolution optical remote sensing images and more than 1 million instances collected from various regions worldwide. It has been annotated with oriented bounding boxes for 5 categories and 37 fine-grained subcategories. We test all models on the official leadboard-v2.0\footnote{\url{https://www.gaofen-challenge.com/benchmark}} and report the mean average precision (mAP) metric score.

\end{enumerate}

\noindent\textbf{Change Detection.} Change detection aims to find pixel-level regional changes via bi-temporal or multi-temporal images. 
Based on \cite{sun2022ringmo}, we incorporate the backbones of different RSFMs into the BIT framework \cite{chen2021remote} to evaluate their performance on the LEVIR-CD dataset. Following \cite{manas2021seasonal,mall2023change}, we use U-Net \cite{ronneberger2015u} as the segmentation head to evaluate the effectiveness of RSFMs on the bi-temporal change detection task using the OSCD dataset with multispectral imagery. Additionally, we employ the DynamicEarthNet-Sentinel2 dataset to assess the performance of the models on the semantic change detection task, maintaining the same setup as segmentation task. Other setting is shown in \cref{singleimple} (iv).

\begin{enumerate}[label=\arabic*)]
\item \textit{LEVIR-CD \cite{chen2020spatial}.} LEVIR-CD is a dataset focused on building change detection, containing of 637 pairs of visible images with a GSD of 0.5m. Each image has a size of 1024$\times$1024 pixels, and the image acquisition time span ranges from 2002 to 2018. The images from different time periods exhibit significant building changes, particularly in the area with rapid population growth. In addition, binary labels (1 indicating change, 0 indicating no change) are provided to indicate the buildings' change status in these bi-temporal images. We report the F1-score on the test set using the same split as \cite{sun2022ringmo}.

\item \textit{OSCD \cite{daudt2018urban}.} This dataset contains 24 pairs of multispectral images obtained from the Sentinel-2 sensor. Following the setup in \cite{manas2021seasonal}, 14 pairs are used for training, and the rest pairs are used for testing. By dividing the original images into non-overlapping patches of size 96$\times$96 pixels, we obtain 827 patches for training and 285 patches for testing.

\item \textit{DynamicEarthNet-Sentinel2 (Dyna.-S2) \cite{toker2022dynamicearthnet}.} The DynamicEarthNet-Sentinel2 dataset can be used to assess the performance of RSFMs on the semantic change detection task as well. Unlike semantic segmentation task, we report the semantic change segmentation (SCS) score, which consists of two components: a class-agnostic binary change score (BC) and a semantic segmentation score among changed pixels (SC). The BC score quantifies the accuracy in identifying changing pixels, while the SC score reflects the ability of methods to accurately predict the category where the changed pixels belong to.
\end{enumerate}

\noindent\textbf{Scene Classification.} We select two commonly-used single-label scene classification datasets (\ie~AID and NWPU-RESISC45 datasets), a multi-label multispectral scene classification dataset (\ie~BigEarthNet-Sentinel2 dataset), and a temporal multispectral scene classification dataset (\ie~fMoW-Sentinel2 dataset). We conduct scene classification experiments using a standard linear classifier. The implementation details are provided in \cref{singleimple} (\texttt{iii}).

\begin{enumerate}[label=\arabic*)]
\item \textit{AID \cite{xia2017aid}.} The AID dataset comprises 10000 images, each with a size of 600$\times$600 pixels and a GSD ranging from 0.5 meters to 8 meters. The images in the dataset are divided into 30 categories, with each category containing approximately 220 to 400 images. In experiments, we use $x\%$ of the data for training and the rest $(1-x \%)$ for testing, following common protocols in \cite{sun2022ringmo, wang2022advancing}, where $x\in\{20, 50\}$.

\item \textit{NWPU-RESISC45 (RESISC-45) \cite{cheng2017remote}.} This dataset includes 31500 images, each with a size of 256$\times$256 pixels and a GSD ranging from 0.5 meters to 30 meters. It is divided into 45 categories, with each category containing 700 images. Similar to previous works \cite{sun2022ringmo, wang2022advancing}, we use 10\% and 20\% of the data for training and the remaining 90\% and 80\% for testing respectively.

\item \textit{BigEarthNet-Sentinel2 (BEN-S2) \cite{sumbul2020bigearthnet}.} The BigEarthNet-Sentinel2 dataset is a large-scale multispectral image dataset used for multi-label land cover scene classification. This dataset consists of a total of 590326 multispectral images, each of which provides multiple land use category annotations. In line with previous studies \cite{manas2021seasonal, wanyan2023dino}, we adopt the new scheme of 19 classes proposed in \cite{sumbul2020bigearthnet}, and exclude approximately 12\% of the patches that are completely masked by seasonal snow, clouds, or cloud shadows in the experiment. In addition, we employ the same data splits as \cite{manas2021seasonal,wanyan2023dino,wang2022ssl4eo}, where 311667 samples are allocated for training and 103944 images are reserved for validation.

\item \textit{fMoW-Sentinel2 (fMoW-S2) \cite{cong2022satmae}.} The fMoW-Sentinel2 dataset is an extension of the fMoW-RGB dataset \cite{christie2018functional}, focusing on temporal multispectral scene classification. For each location, time-series Sentinel-2 images and their corresponding labels are provided. The dataset consists of images with 13 spectral bands provided by Sentinel-2 (B1-12 and B8A), which are captured at different points in time. Some of the time points are the same as the original fMoW-RGB and others are additional points to form a decent time series. In total, the dataset contains 712874 training images, 84939 validation images, and 84966 test images. Following previous works \cite{cong2022satmae,fuller2023croma}, we fine-tune the models using the complete training set and report the Top-1/5 Accuracy metrics on the validation set.

\end{enumerate}

\begin{figure*}[htbp]
	\centering
		\includegraphics[scale=.047]{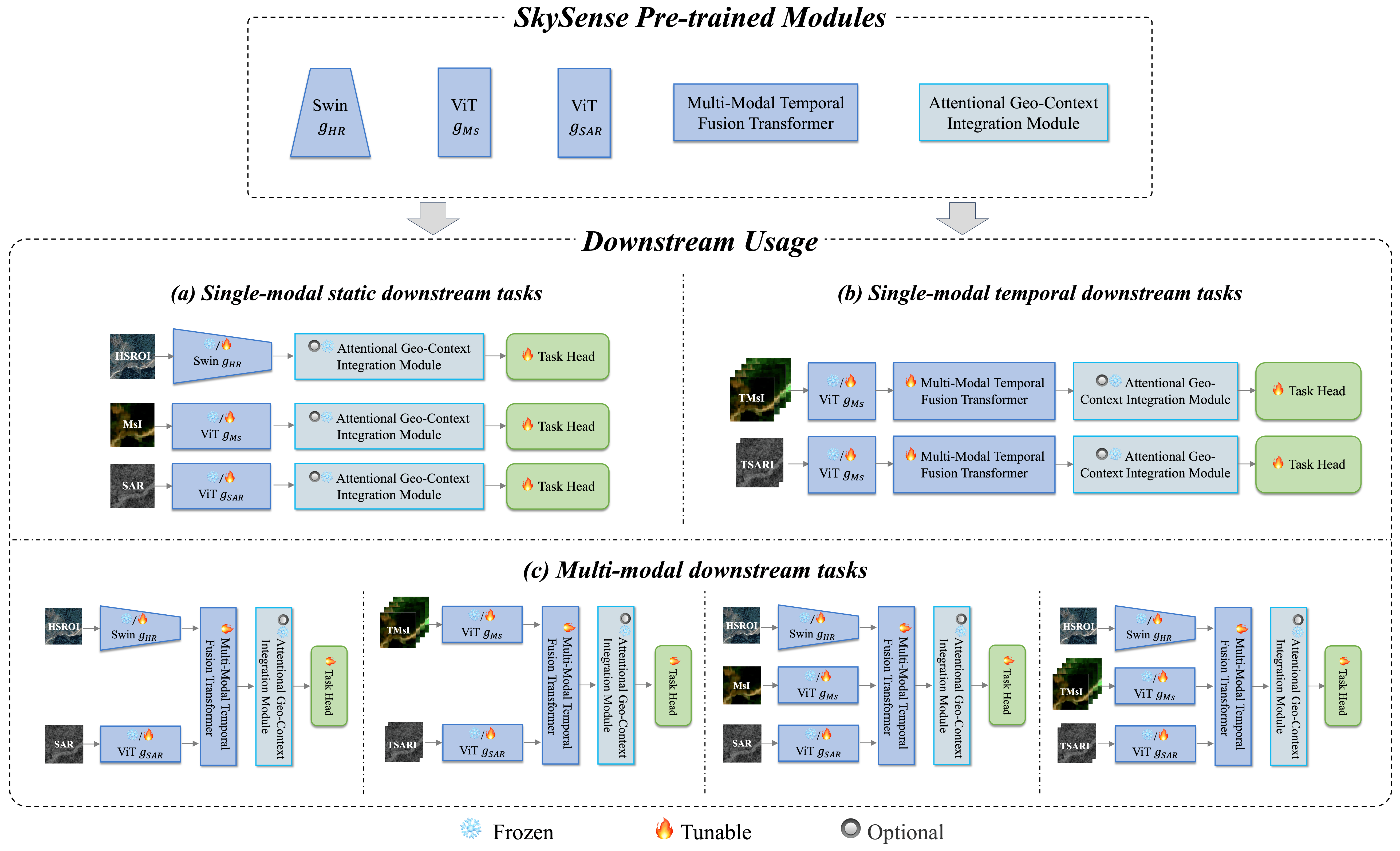}
	    \caption{A detailed illustration on combination of pre-trained modules to accommodate different tasks.}
	\label{fig:downstream}
\end{figure*}

\begin{figure}[htbp]
  \centering
  \includegraphics[width=0.48\textwidth]{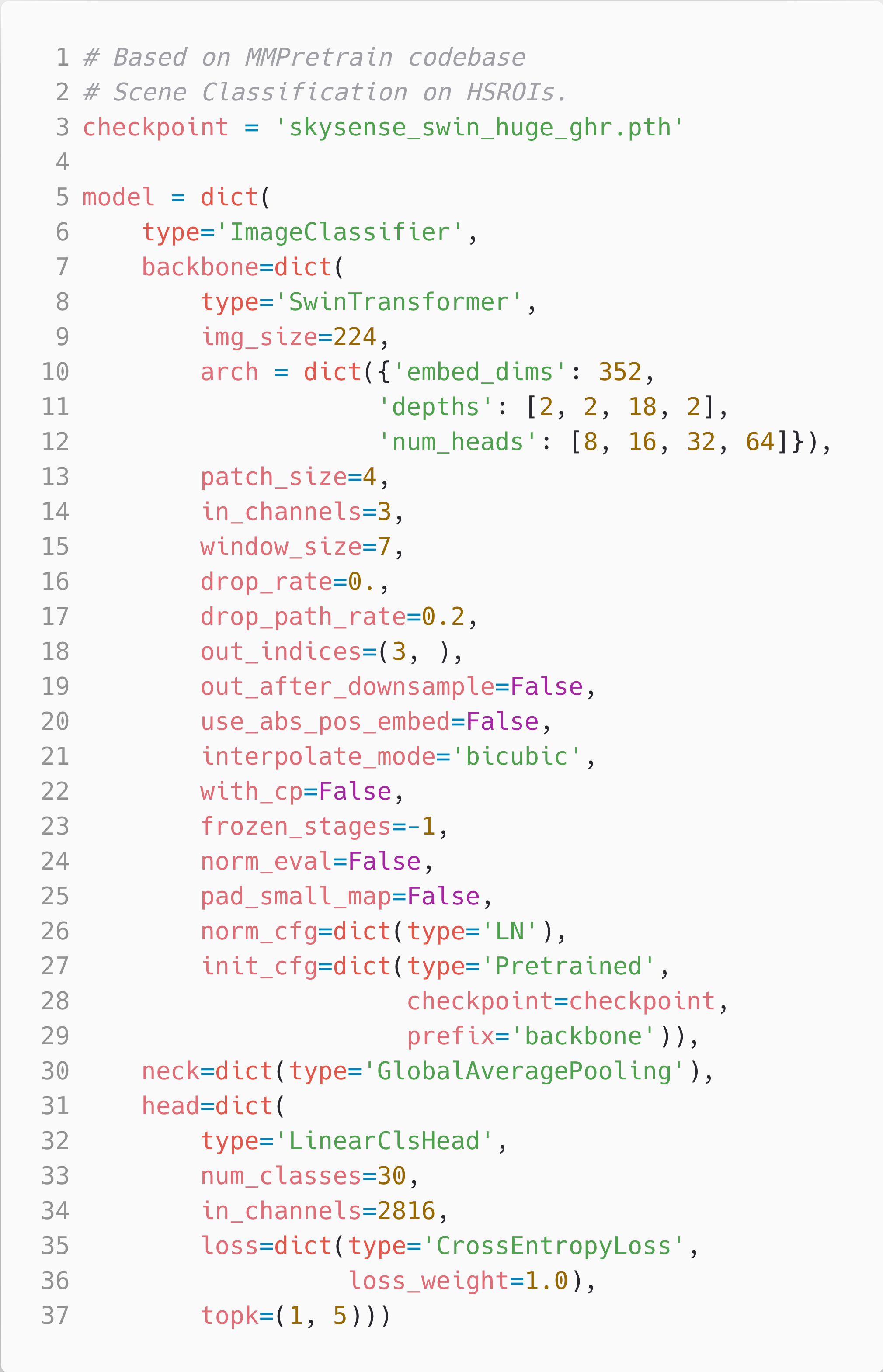}
  \caption{Example configuration for using SkySense pre-trained weights in single-modal scene classification task on HSROIs.}
  \label{fig:code}
\end{figure}

\noindent\textbf{Multi-Modal Semantic Segmentation.} By integrating multi-modal data from different sensors, imaging mechanisms, resolutions, and spectral bands, we can obtain a more diverse and discriminative features. These features enhance the understanding and interpretation of the shape, size, and relationships among ground objects. Thus, we employ the DynamicEarthNet-MM dataset and the PASTIS-MM dataset to evaluate the tasks of Time-insensitive Land Cover Mapping and Time-sensitive Crop Mapping, respectively.

\begin{enumerate}[label=\arabic*)]
\item \textit{DynamicEarthNet-MM (Dyna.-MM) \cite{toker2022dynamicearthnet}.} This dataset contains spatially and temporally aligned multi-modal data, including PlanetFusion imagery (\ie~DynamicEarthNet-PlanetFusion dataset), Sentinel-2 multispectral imagery (\ie~DynamicEarthNet-Sentinel2 dataset), and Sentinel-1 SAR imagery. In case of SAR data, we collect standard-calibrated Sentinel-1 GRD data (VV, VH Polarization) according to the geographical coordinates of the optical imagery, thereby fulfilling the requirements of the multi-modal experiment. To ensure consistency, we employ UperNet as the segmentation head and present the mIoU metric derived from the official Leaderboard-test evaluation hosted online. More implementation details can be seen in \cref{multimodalimple} (\texttt{i}).

\item \textit{PASTIS-MM \cite{garnot2022multi}.} We develop the PASTIS-MM dataset for the task of fine-grained time-sensitive crop mapping. This dataset is an extension of the PASTIS-R dataset \cite{garnot2022multi}, incorporating spatially aligned high-resolution RGB images. It aims to investigate the impact of the combined usage of high-resolution optical imagery, medium-resolution temporal multispectral data, and temporal SAR data in the field of time-sensitive crop mapping. To create the PASTIS-MM dataset, we extract the geo-coordinates and acquisition dates from the image tiles of PASTIS-R dataset. Then we match every image tile with its corresponding static high-resolution optical image, whose GSD is about 0.3 meter. The PASTIS-MM dataset consists of 2433 Sentinel-2 TMsI, each having an image size of 128$\times$128 pixels, 10 spectral bands, and a GSD of 10 meters. It also includes Sentinel-1 GRD SAR images (with VV, VH, and VV/VH channels) and static high-resolution RGB images that we added.
For each image tile, the dataset provides all available Sentinel-2 and Sentinel-1 acquisition data between September 2018 and November 2019, along with the additional high-resolution visible acquisition. Based on statistical analysis, each time series includes approximately 33 to 61 multispectral acquisitions, 70 radar acquisitions, and one high-resolution visible acquisition. The dataset encompasses 18 crop categories and covers a geographical area exceeding 4000 square kilometers. We intend to release this extended dataset publicly to facilitate the advancement of agriculture-vision research. In our experiments, the cloud coverage ratio of the Sentinel-2 images is obtained by utilizing Sentinel Hub's cloud detector\footnote{\url{https://github.com/sentinel-hub/sentinel2-cloud-detector}}. In addition, we use a naive FCN head \cite{long2015fully} and report the Overall Accuracy (OA) from the official five-fold validation on this dataset. More implementation information can be seen in \cref{multimodalimple} (\texttt{ii}).


\end{enumerate}

\noindent\textbf{Multi-Modal Scene Classification.} We further employ the representative BigEarthNet-MM dataset to assess the performance of the Skysense in large-scale scene classification task, considering the integration of optical and SAR data. Detailed implementation details are provided in \cref{multimodalimple} (\texttt{iii}).

\begin{enumerate}[label=\arabic*)]
\item \textit{BigEarthNet-MM (BEN-MM) \cite{Sumbul2021bigearthnet}.} The BigEarthNet-MM dataset expands upon the aforementioned BigEarthNet-Sentinel2 dataset by incorporating corresponding Sentinel-1 SAR data, facilitating the evaluation of multi-modal (optical and SAR) multi-label scene classification task. The BigEarthNet-MM dataset supplements each Sentinel-2 image patch in the BigEarthNet-Sentinel2 dataset with a corresponding preprocessed Sentinel-1 image patch that shares a similar timestamp. Additionally, each Sentinel-1 image patch inherits the annotation information from its corresponding Sentinel-2 image patch. The resulting Sentinel-1 image patches possess a GSD of 10 meters, providing dual-polarization information channels (VV and VH), and are based on interferometric wide-swath mode. Following \cite{wang2023decur,fuller2023croma,wang2022ssl4eo}, we adopt the same data splits as the BigEarthNet-Sentinel2 dataset.
\end{enumerate}

\section{Downstream usage with SkySense pre-trained weights}
\label{sup:sec:example}
Our released SkySense pre-trained weights encompass five pivotal modules (as depicted in the upper section of \cref{fig:downstream}): 
\begin{itemize}

\item The Spatial Encoders $g_{HR}$, $g_{Ms}$, and $g_{SAR}$, which are responsible for extracting representations of corresponding modal images.

\item The Multi-Modal Temporal Fusion Transformer module, designed to integrate multi-modal temporal representations.

\item The Attentional Geo-Context Integration module, which introduces the region-specific prototype set to enhance the features.

\end{itemize}

The aforementioned key components of the pre-trained weights can be used alone or combined with the others flexibly to accommodate the requirements of various downstream tasks. We categorize the downstream tasks into three broad types based on the input data types, and we discuss how to utilize the corresponding pre-trained components to suit each of them respectively.

\begin{enumerate}[label=\arabic*)]
\item \textit{Single-modal static downstream tasks.} As shown in the \cref{fig:downstream} (a), the Spatial Encoders for the corresponding modalities are employed to extract spatial representations, with their parameters being either freezable or tunable. Moreover, if the geographic coordinates corresponding to the images are available, the Attentional Geo-Context Integration module can be involved to attentional integrate pre-trained geo-context information. It is noteworthy that the use of the Attentional Geo-Context Integration module is optional for the user. The output features are fed into task-specific heads for further fine-tuning to obtain the desired outputs.

\item \textit{Single-modal temporal downstream tasks.} Different from \textit{single-modal static downstream tasks}, the parameter-learnable Multi-Modal Temporal Fusion Transformer module is applied after the Spatial Encoder to merge features from temporal sequences, as shown in the \cref{fig:downstream} (b).

\item \textit{Multi-modal downstream tasks.} \cref{fig:downstream} (c) demonstrates examples of various multi-modal data combinations. The respective Spatial Encoders are applied to extract features from the corresponding modalities (static or temporal) data. After concatenation and reshaping, these features are fed into the parameter-learnable Multi-Modal Temporal Fusion Transformer module to obtain a fused representation. The optional Attentional Geo-Context Integration module may be considered to further enhance the features before they are input into various task-specific decoders. This process allows for the effective integration and enhancement of multi-modal data, which is crucial for complex downstream tasks that require a comprehensive understanding of both spatial and temporal cues.

\end{enumerate}

SkySense pre-trained weights exhibits compatibility with the MMCV framework\footnote{\url{https://github.com/open-mmlab/mmcv}} as well as other prevalent codebase repositories (\eg~, torchvision\footnote{\url{https://pytorch.org/vision/stable/index.html}} and TIMM\footnote{\url{https://github.com/huggingface/pytorch-image-models}}), necessitating only rudimentary conversions. Herein, we provide a configuration example for single-modal scene classification based on MMPretrain codebase\footnote{\url{https://github.com/open-mmlab/mmpretrain/tree/mmcls-0.x}}, as illustrate in \cref{fig:code}. More comprehensive usage guidelines can be found within our project repository.




\end{document}